\documentclass{article}

\usepackage{microtype}
\usepackage{graphicx}
\usepackage{subfigure}
\usepackage{booktabs} 

\usepackage{hyperref}
\usepackage{makecell}
\usepackage{gensymb}

\usepackage[accepted]{icml2024}

\usepackage{amsmath}
\usepackage{amssymb}
\usepackage{mathtools}
\usepackage{amsthm}

\usepackage{multicol}
\usepackage{multirow}
\usepackage{textcomp}
\usepackage[capitalize,noabbrev]{cleveref}

\theoremstyle{plain}
\newtheorem{theorem}{Theorem}[section]
\newtheorem{proposition}[theorem]{Proposition}

\theoremstyle{definition}
\newtheorem{definition}[theorem]{Definition}

\theoremstyle{remark}

\newcommand{\ie}{\textit{i}.\textit{e}., }
\usepackage[textsize=tiny]{todonotes}

\icmltitlerunning{Submission and Formatting Instructions for ICML 2024}

\icmltitlerunning{Tilt and Average : Geometric Adjustment of the Last Layer for Recalibration}

\begin{document}

\twocolumn[
\icmltitle{Tilt and Average : Geometric Adjustment of the Last Layer for Recalibration}




\begin{icmlauthorlist}
\icmlauthor{Gyusang Cho}{yyz}
\icmlauthor{Chan-Hyun Youn}{yyz}
\end{icmlauthorlist}

\icmlaffiliation{yyz}{Department of Electrical Engineering, KAIST, Daejeon, Republic of Korea}

\icmlcorrespondingauthor{Gyusang Cho}{cks1463@kaist.ac.kr}

\icmlkeywords{Recalibration, Post-hoc Calibration, Confidence, Uncertainty, Reliability, Interpretability}

\vskip 0.3in
]



\printAffiliationsAndNotice{}  

\begin{abstract}

After the revelation that neural networks tend to produce overconfident predictions, the problem of calibration, which aims to align confidence with accuracy to enhance the reliability of predictions, has gained significant importance. Several solutions based on calibration maps have been proposed to address the problem of recalibrating a trained classifier using additional datasets. In this paper, we offer an algorithm that transforms the weights of the last layer of the classifier, distinct from the calibration-map-based approach. We concentrate on the geometry of the final linear layer, specifically its angular aspect, and adjust the weights of the corresponding layer. We name the method Tilt and Average(\textsc{Tna}), and validate the calibration effect empirically and theoretically. Through this, we demonstrate that our approach, in addition to the existing calibration-map-based techniques, can yield improved calibration performance. Code available : \href{https://github.com/GYYYYYUUUUU/TNA_Angular_Scaling}{\texttt{URL}}.
\end{abstract}

\section{Introduction}

As neural networks demonstrate their powerful performance across various fields, their reliability has become a significant concern. The fact that neural networks are miscalibrated\cite{Guo17On} and are possible to assign high confidence to wrong predictions, is a notable issue in diverse applications, especially in scenarios that are safety-critical(medical usage\cite{Litjens2017Medical}, autonomous-driving\cite{Yurtsever2020Driving}), or high-stake and cost-effective(satelite\cite{moskolai2021satellite}). To resolve this issue, model calibration is studied to reflect reliable confidence estimates to quantify the uncertainty of the prediction.

We target the \textit{recalibration} problem for a model trained on a training set, using an additional dataset. Therefore, the primary focus is on improving the calibration performance of the model while maintaining accuracy as learned during the original training.
The methods introduced in conventional works for recalibration are more focused on fitting a calibration map of the neural network. The calibration map is an additionally designed function that takes the logit of the predicted probability as the input, and outputs calibrated results\citep{Guo17On, zhang2020mix, Tomani2022ParameterizedTS}.

In this paper, we adopt a slightly different perspective, opting to modify the weights of the final linear layer rather than creating a new calibration map. Exploiting the neural network as a feature extractor, the last linear layer transforms high-dimensional deep features into class-specific scores for probability estimation. In this process, we propose techniques that leverage the geometry of the feature space where this transformation occurs. As a result, the proposed algorithm, when combined with traditional calibration-map methods, exhibits superior calibration performance compared to conventional approaches.

In summary, we make the following contributions to this work :
\vspace{-3mm}
\begin{itemize}
\item We propose a recalibration algorithm leveraging the geometry of the last linear layer, deviating from the previously suggested calibration-map approach. Due to the \textbf{orthogonality} of this method, it can seamlessly integrate with existing techniques for recalibration problems, and achieve \textbf{better calibration performance} than using the conventional approach alone.
\item We provide theoretical or experimental background based on the geometric interpretation of the feature space and linear layer. We also verify data efficiency and algorithmic integrity through ablation studies.
\end{itemize}

\begin{figure*}[h]
\label{fig:Overview}
    \centering
    \includegraphics[width=0.95\linewidth]{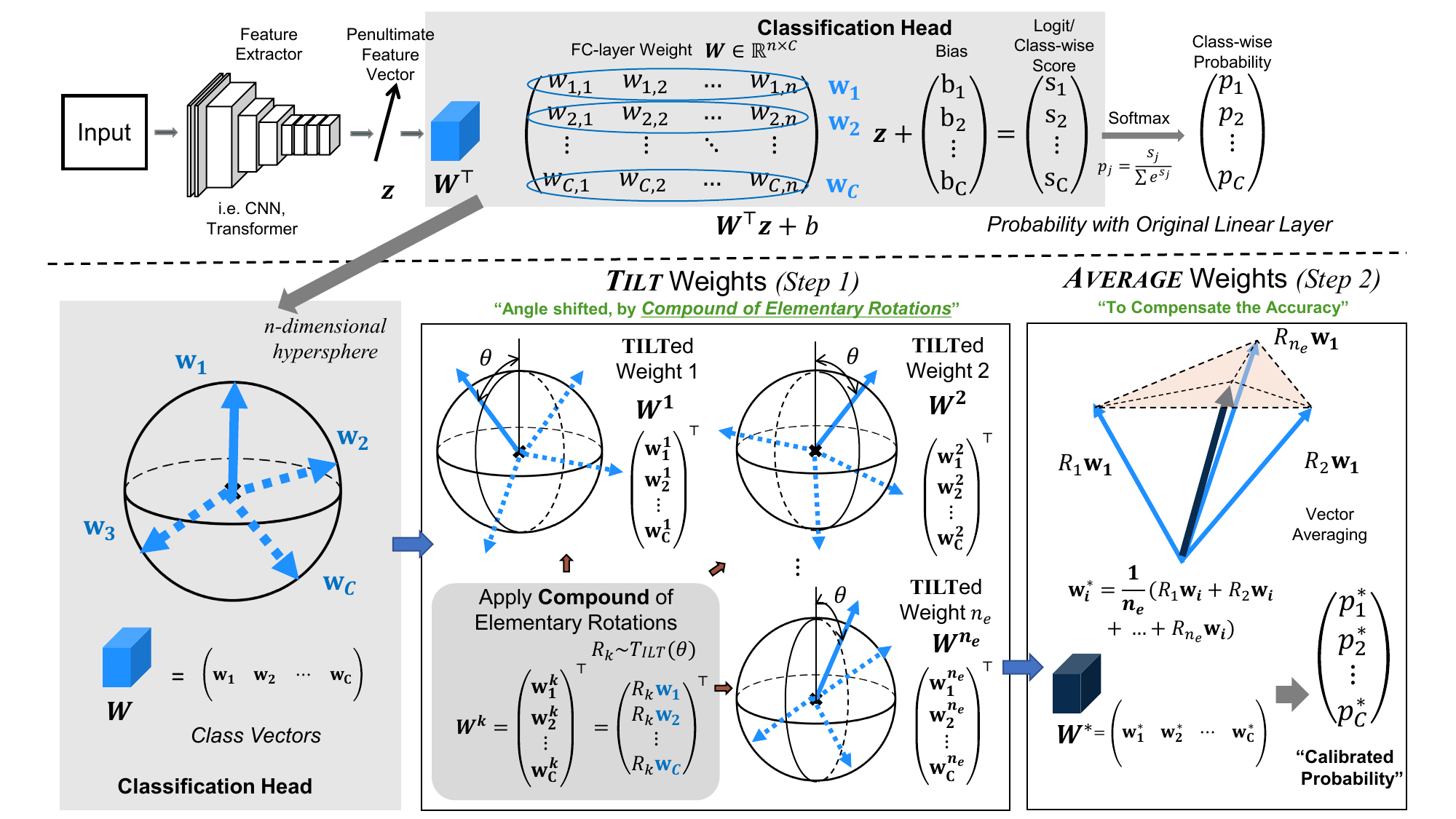}
    \caption{Overview of the proposed algorithm. We take the original weight $W$ of the last linear layer(FC-layer), generate multiple ``\textsc{Tilt}"ed weights $\mathbf{W}^1, \mathbf{W}^2, \cdots \mathbf{W}^{n_e}$ from the original weight with relaxed confidence, and ``\textsc{Average}" the generated weights to compensate possible accuracy loss. The detailed information can be found in \cref{sec:tilt_angles}. }
\end{figure*}

\section{Background \& Related Works}

\subsection{Background}

\textbf{Notation.} We consider the classification task with the labeled dataset. Let the input data $X \in \mathcal{X} $ and the corresponding label $Y \in \mathcal{Y}$ denote the data-label pair, with $C:=|\mathcal{Y}|$ classes. Let the model $M=(f,h)$ trained on the training split of the data, consists of the feature extractor $f: \mathcal{X} \rightarrow \mathbb{R}^n$, and the last classifier function $h:\mathbb{R}^n \rightarrow \mathbb{R}^C$. Then the prediction $\hat{Y}$ and the confidence $\hat{P}$ of a single input $X$ can be written with $\hat{Y}=\arg\max_i{h(f(X))}$, and $\hat{P}=\max_i{\sigma(h(f(X)))}$ for the class $i \in [C]$, where the $\sigma$ is a softmax function.

\textbf{Penultimate Feature, Last Linear Layer, and Confidence.} 
In common cases, the classifier function is a linear layer (fully connected layer) \cite{Kang20Decoupling}. The linear layer applies affine transform on the condensed feature vector $\mathbf{z}=f(X) \in \mathbb{R}^n$ from the feature extractor $f$, \ie, $h(\mathbf{z})=\mathbf{W}^{\top}\mathbf{z}+\mathbf{b}$, where $\mathbf{W} = [\mathbf{w}_1...\mathbf{w}_C]\in \mathbb{R}^{n \times C}$ denotes the weight matrix and the bias vector $\mathbf{b} \in \mathbb{R}^{C}$. We term the vector $\mathbf{z}$ as the \textit{penultimate feature} or \textit{pf}, of which the vector is the output of the penultimate layer of the network. We also designate $\mathbf{w}_i$, the $i$th vector of the classifier weight matrix $\mathbf{W}$, as the $i$th \textit{class vector}, and the weight matrix \textit{original weight}.

Let the logit vector $\mathbf{s} = h(\mathbf{z})$, then the element of the logit vector for each class can be computed with the dot product representation, 
\vspace{-2mm}
\begin{equation}
     s_i = \mathbf{w}_i \cdot \mathbf{z} + b_i = \lVert \mathbf{w}_i\rVert  \lVert\mathbf{z}\rVert \cos \angle(\mathbf{w}_i. \mathbf{z})+b_i, \forall i \in [C]
     \label{eq: class_wise_score}
\end{equation}
Recall that $\hat{P}=\max_i{\sigma(h(\mathbf{z}))}$, the confidence $\hat{P}$ of the sample $X$ with \textit{pf} $z$ can be written as, 
\begin{equation}
     \hat{P} = \max_i{\frac{ \exp({\lVert\mathbf{w}_i\rVert\lVert\mathbf{z}\rVert \cos \angle(\mathbf{w}_i. \mathbf{z})+b_i})}{\Sigma_k{\exp({\lVert\mathbf{w}_k\rVert\lVert\mathbf{z}\rVert \cos \angle(\mathbf{w}_k. \mathbf{z})+b_k})} } },
     \label{eq: confidence}
\end{equation}
where $\angle (u,v)$ denote the angle between two vectors $u,v$. Despite the apparent simplicity of the geometric decomposition in Eq.~\ref{eq: class_wise_score}, \ref{eq: confidence}, significant insights can be derived from this approach. Two properties: 1) the \textbf{norm} (magnitude) of the class vector$\lVert \mathbf{w}_i \rVert$, and 2) the \textbf{angle} between the class vector and the penultimate feature $\angle(\mathbf{w}_i,\mathbf{z})$, are the important factors for determining the confidence. In this paper, we focus on a method that involves adjusting the \textbf{angular property}.

\subsection{Related Works}

\textbf{Calibration, Calibration Error.} \label{sec : cal_NN} Confidence calibration is a problem to correctly predict the probability estimate(i.e. confidence) for a decision that a system has made.
Not only the neural network should make a correct decision, but also it should provide a probability that reflects the ground truth correctness likelihood to make the model interpretable and reliable.  As observed in \cite{Guo17On}, neural networks exhibit poor calibration, typically displaying overconfidence in their predictions\cite{Ovadia2019Can, ashukha2020pitfalls}. 
Formally, the \textit{perfectly calibrated} classifier can be defined as the classifier that outputs $(\hat{Y}, \hat{P})$ to be $\mathbb{P}(\hat{Y} = Y | \hat{P}= p) = p, \forall{p} \in [0,1]$. Following the definition, we evaluate the misalignment, Calibration Error(CE),  $CE=\mathbb{E}_{\hat{P}} \left[ | \mathbb{P}(\hat{Y} = Y | \hat{P}= p) - p | \right]$. The classifier with low $CE$ is considered to be more calibrated.

\textbf{Recalibration,} which is also known as post-hoc calibration, supposes that the classifier trained on the training set is given, and aims to adjust a proper scoring rule to the calculated probability of the network with additional calibration dataset. Most of the work seeks to fit an additional function $f_{cal}:\mathbb{R}^C \rightarrow \mathbb{R}^C$ called \textit{calibration map} which transforms the logit vector $s$ to the calibrated probability vector to yield calibrated confidence estimate $\hat{P}'$. The calibration map $f_{cal}$ is then trained with an additional set of data apart from the training data.  Popular approaches to design $f_{cal}$ includes applying Platt scaling\citep{Platt1999}, temperature scaling\citep{Guo17On} and its variants \citep{zhang2020mix, ding2021local,Tomani2022ParameterizedTS, Joy2023sampleATS}, Dirichlet scaling\citep{Kull2019Beyond}, isotonic regression\citep{Zadrozny2002Transforming}, binning-based approaches\citep{Naeini2015Obtaining, Patel2021Multi}, sample ranking \citep{Rahimi2020Intra, Ma2021a}, non-parametric methods\citep{Wenger2020Nonparametric}, and atypicality-aware map\cite{yuksekgonul2023beyond}.

Orthogonal to the previous findings of designing an advanced calibration map, we take a different perspective and present an algorithm that maps the original weight $\mathbf{W}$ to an alternative weight $\mathbf{W'}$. Treating the weight matrix as a set of \textit{class vectors} $\{w_i\}_{i=1}^{C}$ that determine each element of the logit vector, we seek an alternative set of \textit{class vectors} of $\{w_i '\}_{i=1}^{C}$ that may result in lower calibration error.

\section{Method}

 We aim to design a transformation $Q: \mathbb{R}^n \rightarrow \mathbb{R}^n$, that maps the \textit{class vectors} $\{\mathbf{w}_i\}_{i=1}^C$ to alternative \textit{class vectors} $\{\mathbf{w}_i'\}_{i=1}^C$ that exhibit better calibrated confidence without compromising accuracy. 
 The proposed approach ``Tilt and Average"(\textsc{Tna}) consists of two-fold steps: 1) "\textsc{Tilt}-ing" the class vectors using rotation transformation, and 2) \textsc{Average}-ing weights to supplement the compromised accuracy.

\subsection{``\textsc{Tilt}" the \textit{class vectors} for angle adjustment.}

\label{sec:tilt_angles}
\textbf{Rotation transformation} is a particular type of unitary linear transformation in $n$-dimensional Euclidean space. The key characteristic of a rotation transformation is that the transformed vector undergoes a change in direction without altering its norm, or magnitude. Hence, one might infer that applying a rotation transform to the class vector $\mathbf{w}_i$ could potentially allow for the adjustment of $\angle(\mathbf{w}_i, \mathbf{z})$. Consequently, we elaborate on the algorithm and theoretical background for this purpose as described below.

In this step, we propose an algorithm that can perform rotation transformation to a specified ``intensity". Here, the term ``intense" transform refers to a transformation that introduces significant changes in the angle, while a rotation transform with minimal changes in the angle is considered to have weak intensity. For example, the identity transformation on the class vectors can be regarded as having an intensity of 0, as it introduces no change in angle. Therefore, we first define the concept of ``mean Rotation over Classes" or mRC, which can be considered as a metric for the intensity.

\begin{definition}
\label{def:mRC}
(\textit{mean Rotation over Classes}) 
Given the original weight $\mathbf{W}$ and a rotation matrix $R$, the \textit{mean Rotation over Classes}(\textit{mRC}) is defined as, 
\begin{equation}
mRC(\mathbf{W}, R) = \frac{1}{C}\sum_{i=1}^{C}\arccos{(\frac{\langle \mathbf{w}_i,R\mathbf{w}_i \rangle}{\lVert\mathbf{w}_i\rVert  \lVert R\mathbf{w}_i\rVert})}
\end{equation}
\end{definition}

mRC is a metric quantifying, in the context of predetermined weight matrices and rotation matrices, the typical degree to which each class vector is rotated.

\subsubsection{Generating $n$-dimensional Rotation Matrix with $mRC$}

The main inspiration is from Euler's rotation theorem \citep{EulerRotation} in three-dimensional space that any rotation matrix $R$ can be generated via elementary rotations, $R =  R_x(\alpha)R_y(\beta)R_z(\gamma)$, where $R_x(\alpha)$ states the rotation matrix that rotates of angle $\alpha$ around the X-axis, and vice-versa.
The elementary rotation in dimension $n$, is dubbed Givens rotation \cite{Givens1958}.

We generate a rotation transformation in $n$-dimensional space with a compound of elementary rotations as in Euler's theorem.  Let $n_r$ denote the number of compound of elementary rotations. The elementary rotation matrix $R_k^e(\theta_{t_k})$ ($k \in [n_r]$) can be readily created by using a random angle $\theta_{t_k} = \tau \times \theta_s$, where $\tau \sim Beta(\alpha,\beta)$, and $\theta_s$ is a pre-established angle given as a hyperparameter. We randomly select 2 indexes $k_1, k_2 \in [n]$ from an $n$-dimensional identity matrix. Then we flip the elements of columns and rows $k_1, k_2$ to 2-dimensional rotation matrix, with $r_{k_1 k_1}=\cos{\theta_{t_i}}, r_{k_1 k_2}=-\sin{\theta_{t_i}}, r_{k_2 k_1} =\sin{\theta_{t_i}}, r_{k_2 k_2} =\cos{\theta_{t_i}} $ . We repeat this process $n_r$ times as Eq.~\ref{eq:compoundrot},
\begin{equation}
    R = R_1^e(\theta_{t_1})R_2^e(\theta_{t_2})...R_{n_r}^e(\theta_{t_{n_r}}),
\label{eq:compoundrot}
\end{equation}
which we term the new weight matrix $\mathbf{W^{\textsc{Tilt}}}=R\mathbf{W}$ as \textit{tilted weight}.

\begin{figure}
    \centering
    \includegraphics[width=\linewidth]{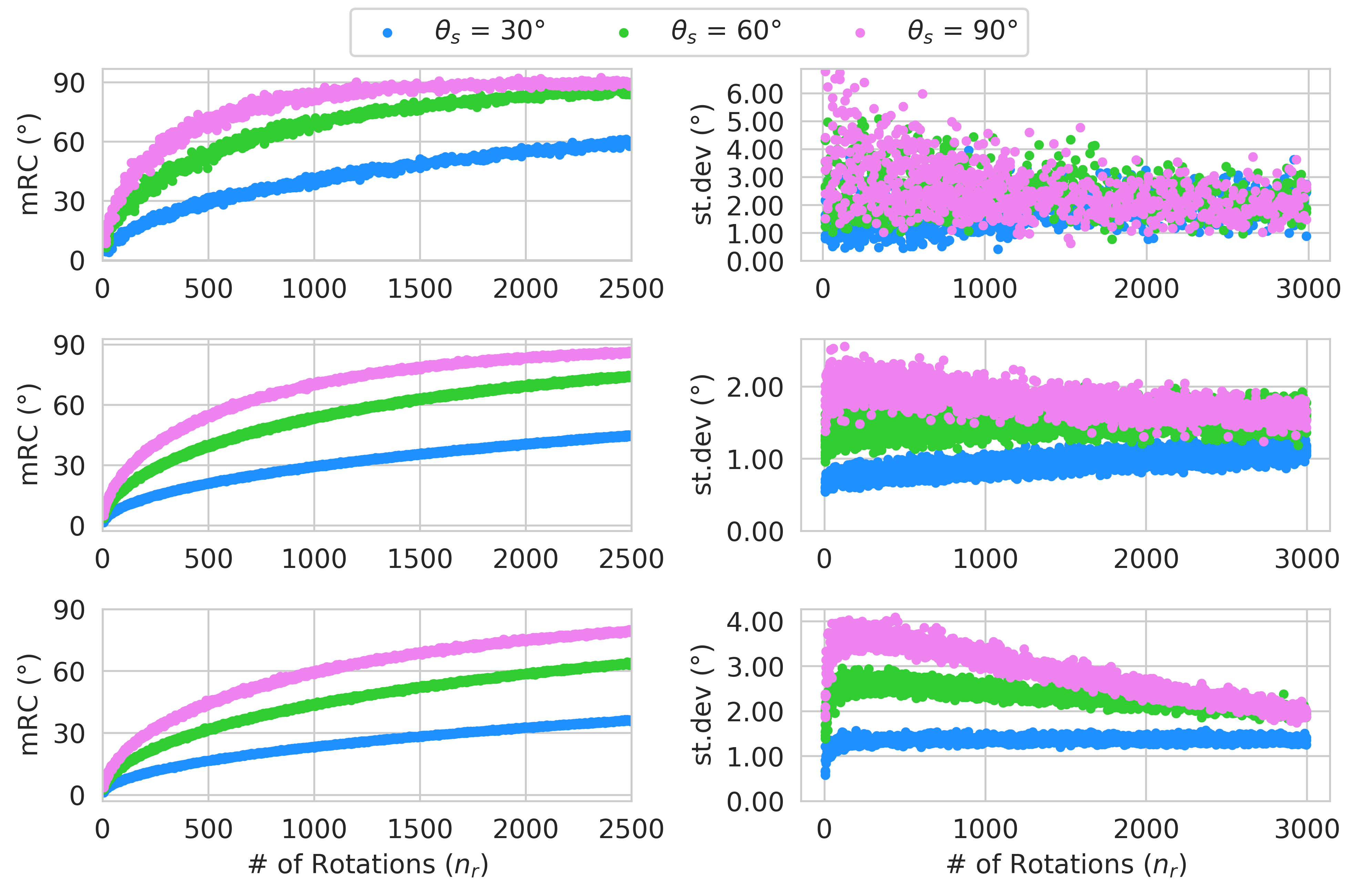}
    \caption{ Plots of \textit{mRC} corresponding to the number of rotations($n_r$) with different values of $\theta_s$. Each point corresponds to a single \textit{tilted weight}. When the number of rotations increases, \textit{mRC} increases as well. Each row states for different dataset-model pair: CIFAR10-WRN28x10(upper), CIFAR100-MobileNetV2(middle), ImageNet-ResNet50(lower). Across datasets and models, the proposed algorithm can generate a rotation matrix with certain mRC, by changing the number of rotations $n_r$.}
    \label{fig:plot_compund_rotations}
\end{figure}

The results of measuring mRC, derived from the rotation matrix R generated by the provided algorithm and the original weight W, are represented in Fig.~\ref{fig:plot_compund_rotations}. For the experiment, we take three benchmark datasets CIFAR10, CIFAR100 \cite{krizhevsky2010cifar}, ImageNet\cite{deng2009imagenet}, with trained weights on the architecture of WideResNet28x10\cite{zagoruyko2016wide}, MobileNetV2\cite{Sandler2018MobileNetV2}, and ResNet50\cite{he2016deep} respectively. We generate 2,000 tilted weights for each dataset. The proposed algorithm can produce a rotation matrix with a specific mRC, demonstrating variations by adjusting the number of rotations $n_r$ across datasets and models. Consequently, we design the algorithm as in Alg.~\ref{alg:tilt}-\textsc{Tilt}. Additionally from the plot, it can be observed that as $n_r$ increases, mRC increases and eventually converges to around 90\textdegree. Applying a transformation with high intensity leads to a decreasing correlation with the original vector, aligning with the near-orthogonal theorem~\cite{blum_hopcroft_kannan_2020}, which states that, with high probability, high-dimensional arbitrary vectors become orthogonal. (Also see \cref{sec: discussion})

\subsubsection{Validating \textsc{Tilt}}

In this section, we theoretically demonstrate that the tilted weight generated by the algorithm presented above allows for the adjustment of angles between class vectors and penultimate features. Additionally, we illustrate the statistical properties associated with this adjustment. With some additional assumptions, we theoretically show the relaxing effect on confidence. Finally, we experimentally confirm that the tilted weight transformed by the algorithm \textsc{Tilt}, indeed has an effect on adjusting angles.

With rotation matrix $R$, we first define the difference after rotation $\Delta_{\mathbf{z}, i} := \angle(R\mathbf{w}_i,\mathbf{z})-\angle(\mathbf{w}_i,\mathbf{z})$, where $\mathbf{z}$ denote the pf.

\begin{theorem}
\label{thm:ndim_mainthm_lem_main}
(Class-wise Effect of \textsc{Tilt}.) Let there be an original weight $\mathbf{W}$, and rotation matrix R. Also, let $\psi_i$ to be $\angle{(\mathbf{w}_i, \mathbf{z})}$. Suppose the rotation matrix $R$ rotates the $i$-th class vector with $\theta$, $\angle{(\mathbf{w}_i, R\mathbf{w}_i)} = \theta$. We further assume that inequality 0\textdegree $< \theta <\psi_i<90$\textdegree holds for penultimate feature $\mathbf{z}$. Lastly, we assume equal probability on the possible rotations of $R$. That is, let $V$ be the set of vectors rotated from a vector $u$ by all possible rotation matrices $R$, that rotates with angle $\theta$, then for $\forall v_1, v_2 \in V$, $\mathbb{P}({Ru = v_1}) =\mathbb{P}({Ru = v_2})$. Let $\mathbb{M}$ be the mode of $\Delta_{\mathbf{z}, i}$. Then the equation below holds, 
\begin{equation}
 \mathbb{M}[\Delta_{\mathbf{z}, i}] = \arccos(\cos\psi_i\cos\theta) - \psi_i.
\end{equation}
\end{theorem}

We provide the proof of Thm.~\ref{thm:ndim_mainthm_lem_main} in the appendix, Sec.~\ref{app: proof}. Building upon the insights derived from the given Thm~\ref{thm:ndim_mainthm_lem_main}, with additional assumptions, one can observe a confidence relaxation effect akin to the proposition as Prop.~\ref{prop: confidence_relaxation}.

\begin{proposition}
\label{prop: confidence_relaxation}

(Confidence Relaxation.) For an input sample $X$ and the corresponding penultimate feature $\mathbf{z}$, we assume that the equalities below hold across all the classes in addition to the assumptions made in Thm.~\ref{thm:ndim_mainthm_lem_main}, $ \forall i\in [C],\angle{(\mathbf{w}_i, R\mathbf{w}_i)} = \theta $ and $\Delta_{\mathbf{z}, i} = \arccos(\cos\psi_i\cos\theta) - \psi_i$, and element of bias vector be $b_{k_1}=b_{k_2}, \forall k_1,k_2\in[C]$.
Then the tilted weight $\mathbf{W}'=R\mathbf{W}$ has a smaller confidence estimate for sample $x$.
\end{proposition}

\begin{proof} Let $\hat{P}$ be the confidence estimate of sample $X$ from original weight $\mathbf{W}$, and $\hat{P'}$ be the confidence estimate of sample $x$ on the tilted weight $\mathbf{W'}$. When applying the assumed conditions to Eq.~\ref{eq: confidence}, then $\hat{P}$ and $\hat{P'}$ can be written as, 
\begin{gather}
     \hat{P_X} = \max_i{\frac{ \exp({\lVert\mathbf{w}_i\rVert\lVert\mathbf{z}\rVert \cos \psi_i)})}{\Sigma_k{\exp({\lVert\mathbf{w}_k\rVert\lVert\mathbf{z}\rVert \cos \psi_k})} } }, \\
     \hat{P}'_X = \max_i{\frac{ \exp({\lVert\mathbf{w}'_i\rVert\lVert\mathbf{z}\rVert \cos \psi_i \cos\theta})}{\Sigma_k{\exp({\lVert\mathbf{w}'_k\rVert\lVert\mathbf{z}\rVert \cos \psi_k \cos \theta})} } },
     \label{eq: confidence_alt}
\end{gather}
    

Recall the fact that rotation transformation does not change the vector norm and the norm of vectors are non-negative, $\hat{P}_X - \hat{P}'_X > 0$ when 0\textdegree $< \theta < 90$\textdegree.

\end{proof}
\begin{figure}[ht]
\begin{center}
\centerline{\includegraphics[width=\columnwidth]{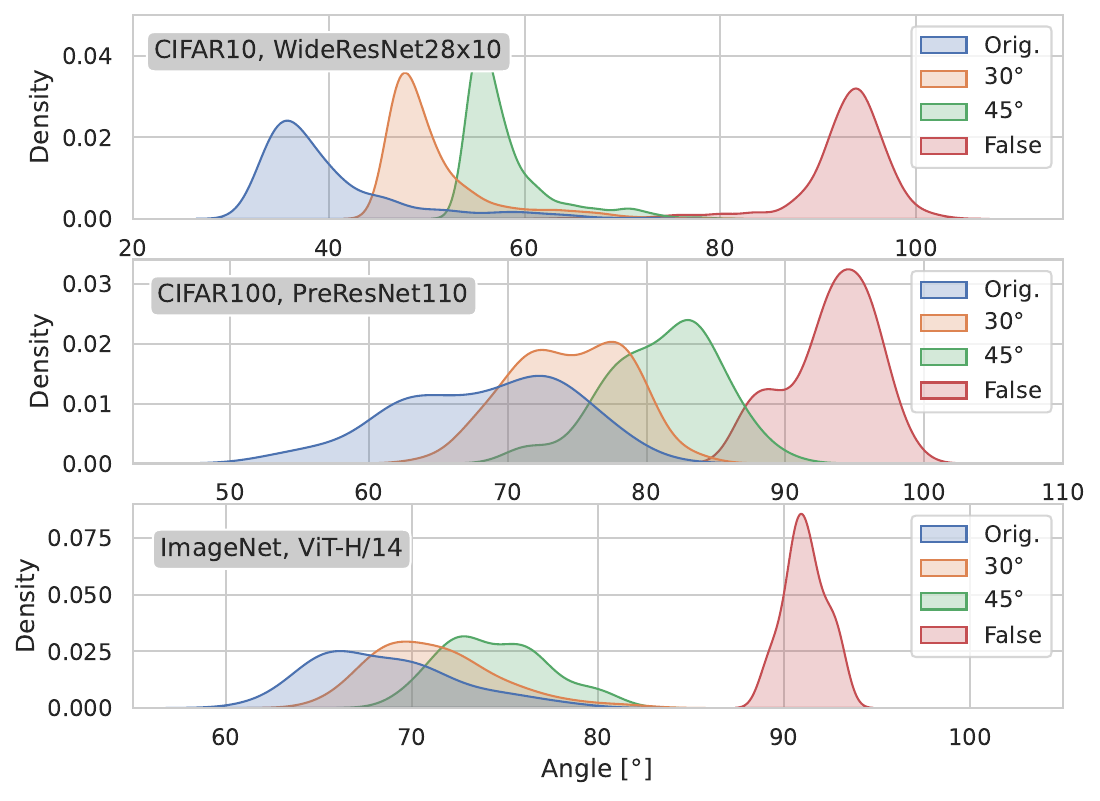}}
\caption{Distribution plot of the data samples with angle between class vector and pf, $\angle(\mathbf{w}_i, \mathbf{z}_x)$, with the corresponding predicted class row vector of the \textbf{original weight} (Orig.) and the \textbf{tilted weights} by depicted angle of $mRC$,  30\textdegree, 45\textdegree. ``False" denotes the angle of $pf$ with the class vector does not correspond to the respective class. As the $mRC$ increases, the angles shift towards 90 \textdegree.  
The distributions of various dataset-architectures can be observed in the appendix, specifically in Fig.~\ref{fig:data_distribution_supp_CF} and ~\ref{fig:data_distribution_supp_IN}. The visuals are enhanced with colors.}
\label{fig:data_distribution}
\end{center}
\end{figure}

The actual distribution of the data when the rotation transformation is applied is illustrated in Fig.~\ref{fig:data_distribution}. This figure depicts the distribution of $\angle(\mathbf{w}_i, \mathbf{z}_x)$.  Formally, we denote data pair as $(x, y) \in \mathcal{X}\times \mathcal{Y}$, penultimate feature as $\mathbf{z}_x$, original weight as $\mathbf{W}$, tilted weight $\mathbf{W}^1= R_1 \mathbf{W}=[\mathbf{w}_1^1 \mathbf{w}_2^1 ...\mathbf{w}_C^1]$ where $mRC(\mathbf{W},R_1) = 30$\textdegree, tilted weight of mRC=45\textdegree as $\mathbf{W}^2= R_2 \mathbf{W}=[\mathbf{w}_1^2 \mathbf{w}_2^2 ...\mathbf{w}_C^2]$ where $mRC(\mathbf{W},R_2) = 45$\textdegree. For distribution depicted \textit{Orig.} we plot $\angle(\mathbf{w}_y, \mathbf{z}_x)$ for the samples. For the 30\textdegree, and 45\textdegree, we plot $\angle(\mathbf{w}_y^1, \mathbf{z}_x)$ and $\angle(\mathbf{w}_y^2, \mathbf{z}_x)$ respectively. For the ``False" plot, we randomly choose a class $y' \neq y$, and compute the $\angle(\mathbf{w}_{y'}, \mathbf{z}_x)$. We randomly sampled 4,000 data from the test dataset. We conducted this experiment on three benchmark datasets, namely CIFAR10, CIFAR100, and ImageNet. For each dataset, we respectively plotted the distribution for WideResNet28x10, PreResNet110, and ViT-H/14 architectures. 
Experimentally, we demonstrate the capability to modify angles using the algorithm depicted in this distribution plot across various datasets and models. Specifically, as $mRC$ increases, it can be observed that the angle distribution gradually shifts towards 90 degrees.

\subsection{"\textsc{Average}" tilted weights for Accuracy Compensation.}

\label{sec:acc_compensation}
As evident from Fig.~\ref{fig:data_distribution}, adjusting angles through the proposed algorithm can lead to overlap with the distribution of the false class, eventually resulting in a lack of separability for the classifier. This, in turn, may lead to a deterioration in the overall performance of the classifier. Therefore, we propose to tune the network by weight averaging to address this issue. We simply apply weight averaging on the multiple tilted weights $\mathbf{W^1},..., \mathbf{W}^{n_e}$ generated on $mRC=\theta$ as in Eq.~\ref{eq: Averaging}, where $n_e$ denotes the number of the tilted weights.

\vspace{-5mm}
\begin{equation}
    \mathbf{W}_{TNA} = \frac{1}{n_e}(\mathbf{W^1}+\mathbf{W^2}+\cdots+ \mathbf{W}^{n_e}),\text{ } 
    \label{eq: Averaging}
\end{equation}
In practice, we generate multiple rotation matrices $R_1, ..., R_{n_e}$ with the same number of $n_r$, and average to get the final transformation matrix $R$ for computational simplicity as in Alg.~\ref{alg:tilt}-\textsc{Average}.

\textbf{Vector Averaging on equally tilted weights.} Weight Averaging averages the \textit{tilted class vectors} generated by equal mRC. This relieves the probability of collapsing in accuracy because weight averaging works as a \textit{arithmetic mean} over $s_i=\mathbf{w_i} \cdot \mathbf{z}$, and \textit{geometric mean} over the $e^{s_i}$, which are the softmax members if ignoring the bias. If the rotation happens to the unwanted direction, averaging weight acts as a geometric mean on the softmax member, and may compensate for the final score through statistical effects.

\begin{figure}
    \centering
    \includegraphics[width=0.85\linewidth]{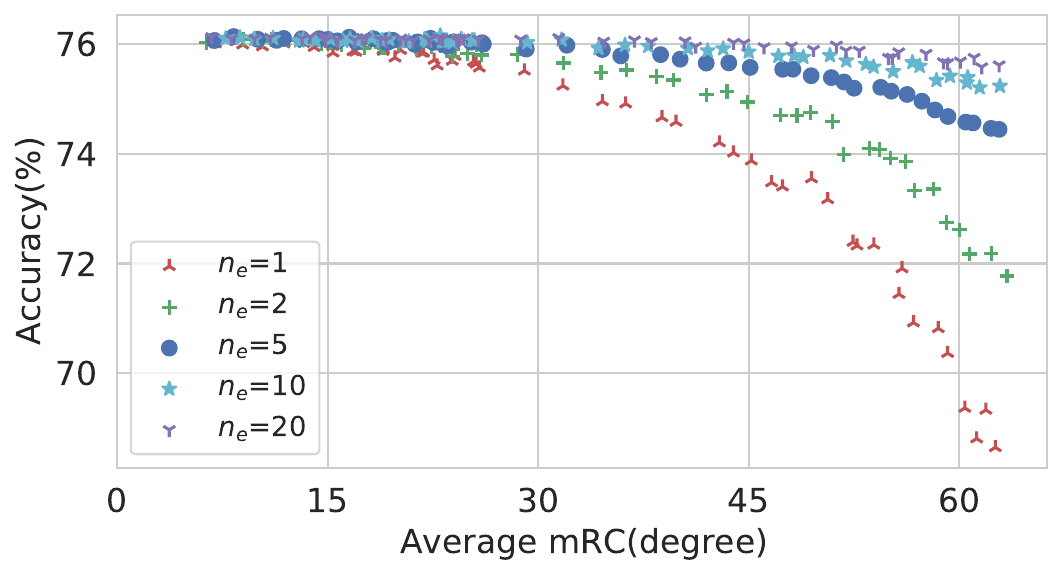}
     \includegraphics[width=0.85\linewidth]{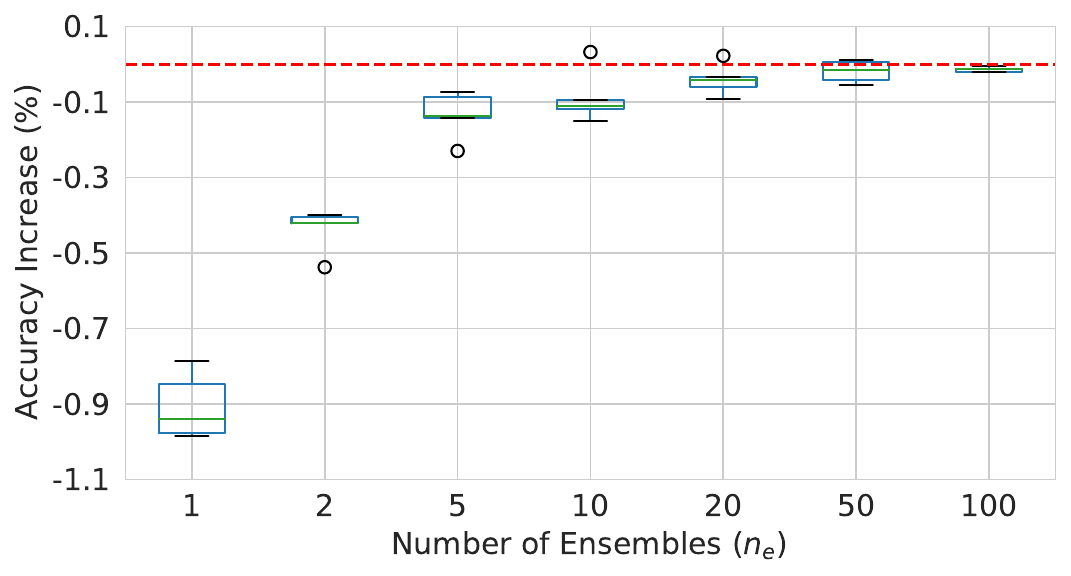}
    \caption{Plots of the accuracy of the ensembled outputs correspond to the number of ensemble members, on the ImageNet dataset. The accuracy is well compensated at most of the angles (upper) and as the $n_e$ increases (lower). The red line indicates the performance of the original weight. The performance is averaged over 10 runs. Extended experiments at appendix Fig.~\ref{fig:accuracy_compensation_supp_CF}}
    \label{fig:accuracy_compensation}
\end{figure}

We illustrate the plot for the interpolated accuracy results through weight averaging in Figure~\ref{fig:accuracy_compensation}. This experiment was conducted on the ImageNet dataset using the ResNet50 architecture. Initially, tilted weights with $mRC=\theta$ were generated, and for each of the generated tilted weights (totaling $n_e$), weight averaging was applied, and the corresponding accuracy was recorded. Each point in the plot represents the performance of the classifier when the averaged weight is applied. The upper plot presents the results of experiments conducted for different mRCs by varying $n_e$ from 1 to 20, while the lower plot records the averaged accuracy for a fixed mRC of 30 degrees over a specified $n_e$ for 1 to 100.
Through this, we observe that as the intensity of the rotation transformation increases, there is a decline in accuracy, but this degradation can be interpolated through weight averaging. In other words, the proposed algorithm effectively interpolates accuracy under varying levels of rotation transformation intensity.

\begin{algorithm}[H]
\caption{\textsc{Tilt And Average (Tna)} }
\label{alg:tilt}
\textbf{Input}: m$\mathcal{RC}$ $\theta^*$, original weight $\mathbf{W}$\\

\textbf{Parameter}:  $n_e$, $n_t$, $\alpha, \beta$, $\theta_s$\\
\textbf{Output}: alternative weight $\mathbf{W}'$ 
\begin{algorithmic} 
 
\STATE (1. (\textsc{Tilt}) Rotation Transform of m$\mathcal{RC}$ $\theta^*$ Generation)
\STATE $R_1 \leftarrow $ Identity Matrix $I_n$, $n_r \leftarrow 0$ 

\WHILE{$mRC (\boldsymbol{W},R_1) \leq \theta^*$}
\STATE $R^t \leftarrow $ Identity Matrix $I_n$
\STATE $ \tau \sim Beta(\alpha, \beta)$ ,  $\theta_t = \theta_s \times \tau$, $k_1, k_2 \in [n]$  
\STATE $R^t_{k_1 k_1} = \cos{\theta_t}$, $R^t_{k_1 k_2} = -\sin{\theta_t}$ 
\STATE $R^t_{k_2 k_1} = \sin{\theta_t}$, $R^t_{k_2 k_2} = \cos{\theta_t}$
\STATE $R_1 \leftarrow R_1R^t$
\STATE $ n_r \leftarrow n_r + n_t$ 
\ENDWHILE

\STATE (2. (\textsc{Average}) Compensate Accuracy)
\STATE Repeat Procedure (1) $n_e$ times and get $R_1, R_2, ..., R_{n_e}$
\STATE $R \leftarrow \frac{1}{n_e}(R_1 + R_2 + ... + R_{n_e})$
\STATE $\mathbf{W}' \leftarrow R\mathbf{W}$
\STATE \textbf{return} $\mathbf{W}'$
\end{algorithmic}
\label{alg:randomrot}
\end{algorithm}
 \vspace{-7mm}

\section{Experiments}

\subsection{Evaluation Metrics}

We report the Accuracy, ECE, and AdaECE in $\%$ in table\ref{tab:hardness}. In practice, the Calibration Error is intractable since the joint distribution $\mathbb{P}(X,Y)$ is unknown. To evaluate the calibration performance of the classifiers, we estimate Calibration Error(noted in \cref{sec : cal_NN}) by 2 different schemes, \textbf{ECE} and \textbf{AdaECE}. We also display the accuracy as well, to demonstrate the accuracy change after applying \textsc{Tna}.

\textbf{Expected Calibration Error}, or \textbf{ECE}, estimates Calibration Error by \textit{equal-interval-binning} scheme. Suppose there exists $B$ bins of $|D|$ data samples, where each bin $B_j, j \in \{ 0,1,...,B-1 \} $ consists of the samples with confidence corresponding to the interval $I_j=\left(\frac{j}{B}, \frac{j+1}{B} \right]$. Then, $\textbf{ECE} = \Sigma_{j}\frac{|B_j|}{|D|}|acc(B_j)-conf(B_j)|,$ where $acc(B_j)$ is the accuracy of the samples from $B_j$, and $conf(B_j)$ is the average confidence of the samples from $B_j$ \cite{Naeini2015Obtaining}.

\textbf{Adaptive Expected Calibration Error}, or \textbf{AdaECE}, estimates Calibration Error by \textit{uniform-mass-binning} scheme, with identical number of samples for each bin\cite{mukhoti2020calibrating}. In other words, $\textbf{AdaECE}=\Sigma_{j}\frac{|B_j|}{|D|}|acc(B_j)-conf(B_j)|$ where each bin $B_j$ is set to include $\frac{|D|}{B}$ samples with the ascending order of the confidence value.

\subsection{Experimental Details}
\textbf{Datasets.} We report the results of the proposed method on 4 different image classification datasets. The number at the end represents the number of samples used for training/calibration/test split. The number of splits for each data partition adheres to the methodology outlined in the work of \citet{ashukha2020pitfalls}.
\begin{itemize}
\item CIFAR10, CIFAR100 \cite{krizhevsky2010cifar}: Tiny images from web. 10 and 100 classes respectively. 50,000/5,000/5,000
\item ImageNet-1k \cite{deng2009imagenet} : Image of natural objects, 1000 classes, 1.2M/12,500/37,500
\item SVHN \cite{netzer2011reading}: Street View House Numbers, 10 classes of digits, 73,257/5,000/19,032. Results in the appendix, Tab.~\ref{tab:SVHN results}.
\end{itemize}

\begin{table*}
    \centering 
     \caption{Main results displaying changes after \textsc{Tna}, in accuracy and calibration performance. The experiments were conducted on CIFAR10, CIFAR100, and ImageNet datasets, evaluating four different model architectures for each dataset. In addition to applying five recalibration techniques, including the uncalibrated model as a baseline, we evaluate \textsc{Tna}(sparse) and \textsc{Tna}(comp.), which involve two stages of optimization and explore the entire search space, respectively. The results are displayed as averages over five runs. In a specific dataset-model combination, indicate the technique that achieved the best results among all methods in \textbf{bold}. Lower ECE, AdaECE, the better.}\label{tab:hardness}
    \scalebox{0.65}{
    	\begin{tabular}{c|ccc|ccc|ccc|ccc}
    		\toprule

       \multicolumn{1}{c}{} & \multicolumn{3}{c}{\textbf{CIFAR10-WideResNet28x10}} & \multicolumn{3}{c}{\textbf{CIFAR10-MobileNetV2}} & \multicolumn{3}{c}{\textbf{CIFAR10-PreResNet110}}  & \multicolumn{3}{c}{\textbf{CIFAR10-GoogleNet}}  \\
                \midrule
    		\textbf{Methods} & \textbf{Acc.} ($\uparrow$) &  \textbf{ECE} ($\downarrow$) & \textbf{AdaECE} ($\downarrow$) & \textbf{Acc.} ($\uparrow$) &  \textbf{ECE} ($\downarrow$) & \textbf{AdaECE} ($\downarrow$) & \textbf{Acc.} ($\uparrow$) &  \textbf{ECE} ($\downarrow$) & \textbf{AdaECE} ($\downarrow$) & \textbf{Acc.} ($\uparrow$) &  \textbf{ECE} ($\downarrow$) & \textbf{AdaECE} ($\downarrow$) \\\cmidrule(lr){1-13}
    	\multirow{1}{*}{None} & $96.37_{.15}$ & $1.66_{.1}$ & $1.49_{.09}$ & $92.51_{.26}$ & $3.49_{.16}$ & $3.33_{.24}$ & $95.28_{.15}$ & $2.67_{.13}$ & $2.62_{.14}$ & $95.16_{.22}$ & $2.82_{.19}$ & $2.75_{.21}$  \\ 
            \multirow{1}{*}{\textbf{+\textsc{Tna}}} & $96.37_{.16}$ & \textbf{0.66$_{.14}$} & $0.59_{.1}$ & $92.55_{.25}$ & \textbf{0.78}$_{.17}$ & $0.65_{.24}$ & $95.28_{.11}$ & $0.76_{.03}$ & $0.4_{.04}$ &  $95.17_{.21} $& $1.57_{.18} $& $1.59_{.21} $ \\ \cmidrule(lr){1-13}
            
             \multirow{1}{*}{IROvA} & $96.29_{.19}$ & $0.83_{.29}$ & $0.56_{.11}$ & $92.39_{.24}$ & $1.24_{.15}$ & $0.92_{.29}$ & $95.16_{.09}$ & $0.88_{.12}$ & $0.73_{.11}$ & $95.11_{.18}$ & $1.01_{.25}$ & $0.96_{.08}$ \\
            \multirow{1}{*}{\textbf{+\textsc{Tna}(sparse)}} &$96.32_{.16}$ & $0.83_{.23}$ & $0.55_{.24}$ & $92.39_{.29}$ & $1.22_{.13}$ & $0.87_{.32}$ & $95.31_{.04}$ & $0.74_{.08}$ & $0.4_{.19}$ & $95.07_{.16} $& $0.96_{.26} $& \textbf{0.85}$_{.09} $ \\ 
            \multirow{1}{*}{\textbf{+\textsc{Tna}(comp.)}} & $96.32_{.13}$ & $0.8_{.2}$ & $0.62_{.18}$ & $92.5_{.19}$ & $1.18_{.22}$ & $0.89_{.2}$ & $95.31_{.04}$ & $0.74_{.08}$ & $0.4_{.19}$ & $95.07_{.16} $& $0.96_{.26} $& \textbf{0.85}$_{.09} $ \\ \cmidrule(lr){1-13}
            
            \multirow{1}{*}{TS} & $96.37_{.15}$ & $0.84_{.05}$ & $0.71_{.22}$ & $92.51_{.26}$ & $1.06_{.19}$ & $0.88_{.27}$ & $95.28_{.15}$ & $0.73_{.09}$ & $0.65_{.12}$ & $95.16_{.22}$ & $2.14_{.13}$ & $1.63_{.3}$  \\ 
            \multirow{1}{*}{\textbf{+\textsc{Tna}(sparse)}} & $96.37_{.16}$ & $0.78_{.13}$ & $0.75_{.2}$ & $92.55_{.25}$ & $0.81_{.12}$ & $0.72_{.25}$ & $95.28_{.11}$ & $0.68_{.09}$ & $0.41_{.03}$ & $95.17_{.21} $& $2.09_{.16} $& $1.64_{.31} $ \\ 
            \multirow{1}{*}{\textbf{+\textsc{Tna}(comp.)}} & $96.39_{.12}$ & $0.71_{.14}$ & $0.53_{.06}$ & $92.54_{.19}$ & $0.79_{.09}$ & $0.76_{.19}$ & $95.13_{.18}$ & $0.65_{.11}$ & $0.45_{.11}$ & $95.08_{.16}$ & $1.96_{.1}$ & $1.58_{.13}$  \\ \cmidrule(lr){1-13}
            
            \multirow{1}{*}{ETS} & $96.37_{.15}$ & $0.84_{.06}$ & $0.79_{.19}$ & $92.51_{.26}$ & ${0.85}_{.19}$ & ${0.67}_{.24}$ & $95.28_{.15}$ & $0.67_{.1}$ & $0.5_{.12}$ & $95.16_{.22}$ & $2.14_{.13}$ & $1.63_{.3}$ \\
            \multirow{1}{*}{\textbf{+\textsc{Tna}(sparse)}} & $96.37_{.16}$ & $0.78_{.14}$ & $0.72_{.17}$ & $92.55_{.25}$ & $0.83_{.09}$ & $0.71_{.26}$ & $95.28_{.11}$ & $0.69_{.08}$ & $0.42_{.0}$ & $95.17_{.21} $& $2.09_{.16} $& $1.64_{.31} $ \\ 
            \multirow{1}{*}{\textbf{+\textsc{Tna}(comp.)}} & $96.39_{.12}$ & $0.71_{.14}$ & $0.53_{.06}$ & $92.56_{.2}$ & $0.79_{.1}$ & $0.82_{.25}$ & $94.66_{.18}$ & $0.65_{.11}$ & $0.45_{.11}$ & $95.08_{.16}$ & $1.96_{.1}$ & $1.58_{.13}$   \\ \cmidrule(lr){1-13}
            
            \multirow{1}{*}{AAR} & $96.33_{.2}$ & ${0.75}_{.08}$ & ${0.45}_{.09}$ & $92.38_{.32}$ & $0.91_{.07}$ & $e{0.67}_{.1}$ & $95.17_{.08}$ & $0.59_{.1}$ & $0.45_{.05}$ & $95.14_{.24}$ & $1.04_{.12}$ & $0.95_{.19}$  \\
            \multirow{1}{*}{\textbf{+\textsc{Tna}(sparse)}} & $96.35_{.19}$ & $0.72_{.05}$ & $0.45_{.08}$ & $92.37_{.33}$ & $0.83_{.14}$ & \textbf{0.64}$_{.12}$ & $95.3_{.13}$ & $0.59_{.18}$ & \textbf{0.34}$_{.13}$ & $95.12_{.26} $& $0.98_{.17} $& $0.95_{.22} $\\ 
            \multirow{1}{*}{\textbf{+\textsc{Tna}(comp.)}} & $96.41_{.18}$ & $0.72_{.23}$ & \textbf{0.44}$_{.19}$ & $92.37_{.33}$ & $0.83_{.14}$ & \textbf{0.64}$_{.12}$ & $94.99_{.1}$ & \textbf{0.56}$_{.13}$ & \textbf{0.34}$_{.08}$ & $95.12_{.22}$ & \textbf{0.9}$_{.01}$ & $0.95_{.2}$  \\
            \midrule
            \multicolumn{1}{c}{} & \multicolumn{3}{c}{\textbf{CIFAR100-WideResNet28x10}} & \multicolumn{3}{c}{\textbf{CIFAR100-MobileNetV2}} & \multicolumn{3}{c}{\textbf{CIFAR100-PreResNet110}}  & \multicolumn{3}{c}{\textbf{CIFAR100-GoogleNet}}  \\
                \midrule
                \textbf{Methods} & \textbf{Acc.} ($\uparrow$) &  \textbf{ECE} ($\downarrow$) & \textbf{AdaECE} ($\downarrow$) & \textbf{Acc.} ($\uparrow$) &  \textbf{ECE} ($\downarrow$) & \textbf{AdaECE} ($\downarrow$) & \textbf{Acc.} ($\uparrow$) &  \textbf{ECE} ($\downarrow$) & \textbf{AdaECE} ($\downarrow$) & \textbf{Acc.} ($\uparrow$) &  \textbf{ECE} ($\downarrow$) & \textbf{AdaECE} ($\downarrow$) \\\cmidrule(lr){1-13}
    	\multirow{1}{*}{None} & $80.42_{.47}$ & $5.91_{.3}$ & $5.77_{.4}$ & $72.79_{.14}$ & $10.03_{.19}$ & $9.97_{.18}$ & $77.69_{.35}$ & $10.62_{.33}$ & $10.58_{.34}$ & $79.42_{.26}$ & $6.45_{.2}$ & $6.3_{.19}$   \\ 
            \multirow{1}{*}{\textbf{+\textsc{Tna}}} &$80.17_{.53}$ & $4.11_{.19}$ & $3.92_{.15}$ & $72.71_{.13}$ & $1.54_{.09}$ & $1.52_{.19}$ & $77.15_{.34}$ & $2.97_{.25}$ & $2.84_{.29}$ & $79.36_{.25} $& $3.86_{.2} $& $3.86_{.16} $  \\ \cmidrule(lr){1-13}
            
            \multirow{1}{*}{IROvA} & $80.69_{.36}$ & $4.71_{.14}$ & $4.27_{.15}$ & $72.15_{.22}$ & $4.81_{.28}$ & $4.79_{.19}$ & $76.95_{.25}$ & $3.7_{.17}$ & $4.7_{.49}$ & $78.92_{.27}$ & $3.73_{.35}$ & $4.68_{.17}$   \\
            \multirow{1}{*}{\textbf{+\textsc{Tna}(sparse)}} & $80.4_{.43}$ & $4.15_{.26}$ & $3.36_{.35}$ & $72.01_{.26}$ & $4.72_{.29}$ & $4.17_{.25}$ & $76.51_{.22}$ & $3.91_{.2}$ & $3.65_{.61}$ & $78.88_{.35} $& $3.66_{.35} $& \textbf{3.23}$_{.17} $  \\ 
            \multirow{1}{*}{\textbf{+\textsc{Tna}(comp.)}} & $81.11_{.3}$ & $3.89_{.11}$ & $3.35_{.14}$ & $72.45_{.36}$ & $4.32_{.14}$ & $3.91_{.38}$ & $76.93_{.3}$ & $3.66_{.49}$ & $3.6_{.47}$ & $78.88_{.35} $& $3.66_{.35} $& \textbf{3.23}$_{.17} $  \\ 
            \cmidrule(lr){1-13}

            \multirow{1}{*}{TS} & $80.42_{.47}$ & $4.69_{.22}$ & $4.55_{.26}$ & $72.79_{.14}$ & $1.82_{.33}$ & $1.78_{.11}$ & $77.69_{.35}$ & $3.21_{.17}$ & $3.13_{.17}$ &$79.42_{.26}$ & $4.27_{.12}$ & $4.28_{.15}$ \\ 
            \multirow{1}{*}{\textbf{+\textsc{Tna}(sparse)}} & $80.17_{.23}$ & $4.55_{.24}$ & $4.28_{.14}$ & $72.71_{.13}$ & $1.41_{.15}$ & $1.34_{.11}$ & $77.15_{.34}$ & $3.16_{.26}$ & $3.03_{.35}$ & $79.36_{.25} $& $4.23_{.14} $& $4.13_{.18} $  \\ 
            \multirow{1}{*}{\textbf{+\textsc{Tna}(comp.)}} & $80.17_{.53}$ & $4.55_{.24}$ & $4.28_{.14}$ & $73.13_{.2}$ & $1.41_{.33}$ & $1.24_{.33}$ & $75.64_{.64}$ & $3.01_{.15}$ & $2.97_{.14}$ & $79.36_{.25} $& $4.23_{.14} $& $4.13_{.18} $ \\ 
            \cmidrule(lr){1-13}
            
            \multirow{1}{*}{ETS} & $80.42_{.47}$ & $3.47_{.28}$ & $3.88_{.21}$ & $72.79_{.14}$ & $1.63_{.39}$ & $1.56_{.12}$ & $77.97_{.44}$ & $2.38_{.37}$ & $2.6_{.33}$ & $79.42_{.26}$ & $3.25_{.19}$ & $3.4_{.22}$ \\
            \multirow{1}{*}{\textbf{+\textsc{Tna}(sparse)}} & $80.17_{.53}$ & $3.35_{.26}$ & $3.89_{.22}$ & $72.71_{.13}$ & \textbf{1.22}$_{.18}$ & $1.32_{.03}$ & $77.15_{.34}$ & $2.19_{.28} $& $2.49_{.35} $&  $79.36_{.25} $& \textbf{2.86}$_{.2} $& $3.35_{.26} $ \\ 
            \multirow{1}{*}{\textbf{+\textsc{Tna}(comp.)}} & $80.76_{.27}$ & \textbf{3.06}$_{.16}$ & $3.52_{.17}$ & $73.13_{.2}$ & $1.27_{.32}$ & \textbf{1.18}$_{.29}$ & $75.74_{.72}$ & \textbf{2.13}$_{.27}$ & \textbf{2.4}$_{.37}$ & $79.36_{.25} $& \textbf{2.86}$_{.2} $& $3.35_{.26}$ \\ \cmidrule(lr){1-13}
            
            \multirow{1}{*}{AAR} & $81.06_{.14}$ & $3.4_{.38}$ & $3.19_{.33}$ & $72.53_{.17}$ & $2.06_{.4}$ & $1.82_{.44}$ & $77.37_{.37}$ & $3.27_{.12}$ & $3.23_{.21}$ & $79.28_{.3}$ & $4.46_{.31}$ & $4.28_{.3}$ \\
            \multirow{1}{*}{\textbf{+\textsc{Tna}(sparse)}} &$80.98_{.22}$ & $3.33_{.4}$ & \textbf{3.15}$_{.33}$ & $72.39_{.26}$ & $1.95_{.39}$ & $1.85_{.33}$ & $77.15_{.47}$ & $3.36_{.22}$ & $3.07_{.18}$ & $79.38_{.29} $& $4.45_{.23} $& $4.34_{.28} $ \\
            \multirow{1}{*}{\textbf{+\textsc{Tna}(comp.)}} &$80.98_{.22}$ & $3.33_{.4}$ & \textbf{3.15}$_{.33}$ & $73.09_{.26}$ & $1.52_{.17}$ & $1.47_{.27}$ &  $76.83_{.42}$ & $3.15_{.41}$ & $2.87_{.43}$ & $78.91_{.18}$ & $4.45_{.05}$ & $4.27_{.25}$ \\ 
            \midrule
            \multicolumn{1}{c}{} & \multicolumn{3}{c}{\textbf{ImageNet-ResNet50}} & \multicolumn{3}{c}{\textbf{ImageNet-DenseNet169}} & \multicolumn{3}{c}{\textbf{ImageNet-ViT-L/16}}  & \multicolumn{3}{c}{\textbf{ImageNet-ViT-H/14}} \\
       \midrule
       \textbf{Methods} & \textbf{Acc.} ($\uparrow$) &  \textbf{ECE} ($\downarrow$) & \textbf{AdaECE} ($\downarrow$) & \textbf{Acc.} ($\uparrow$) &  \textbf{ECE} ($\downarrow$) & \textbf{AdaECE} ($\downarrow$) & \textbf{Acc.} ($\uparrow$) &  \textbf{ECE} ($\downarrow$) & \textbf{AdaECE} ($\downarrow$) & \textbf{Acc.} ($\uparrow$) &  \textbf{ECE} ($\downarrow$) & \textbf{AdaECE} ($\downarrow$) \\\cmidrule(lr){1-13}
    	\multirow{1}{*}{None} & $76.17_{.05}$ & $3.83_{.09}$ & $3.73_{.1}$ & $75.63_{.11}$ & $5.43_{.08}$ & $5.43_{.08}$ & $84.35_{.06}$ & $1.81_{.07}$ & $1.8_{.09}$ & $85.59_{.05}$ & $1.87_{.07}$ & $1.77_{.07}$   \\ 
            \multirow{1}{*}{\textbf{+\textsc{Tna}}} &$76.11_{.06}$ & $1.73_{.06}$ & $1.76_{.08}$ & $75.46_{.11} $& $1.63_{.14} $& $1.66_{.14} $& $84.34_{.07} $& $1.11_{.05} $& $1.11_{.05} $&  $85.5_{.03} $& $1.05_{.01} $& $1.06_{.03} $  
\\ \cmidrule(lr){1-13}
             \multirow{1}{*}{IROvA} & $74.96_{.07}$ & $6.23_{.37}$ & $5.63_{.29}$ & $74.82_{.13}$ & $6.19_{.77}$ & $5.72_{.54}$ &$83.5_{.1}$ & $5.35_{.29}$ & $4.27_{.14}$ & 
 $84.75_{.03}$ & $5.3_{.16}$ & $4.31_{.3}$  \\
            \multirow{1}{*}{\textbf{+\textsc{Tna}(sparse)}} &$74.87_{.05}$ & $6.24_{.19}$ & $4.58_{.33}$ & $74.49_{.23} $& $6.1_{.72} $& $4.71_{.67} $& $83.41_{.11} $& $5.32_{.29} $& $4.05_{.12} $& $84.54_{.0} $& $5.24_{.2} $& $4.0_{.24} $ \\ 
            \multirow{1}{*}{\textbf{+\textsc{Tna}(comp.)}} & $74.98_{.03}$ & $6.03_{.11}$ & $4.55_{.21}$  & $74.41_{.13}$ & $6.05_{.61}$ & $4.7_{.56}$ & $83.72_{.08}$ & $5.16_{.11}$ & $4.05_{.01}$ & $84.43_{.13}$ & $5.17_{.11}$ & $3.9_{.16}$ \\ \cmidrule(lr){1-13}
            
            \multirow{1}{*}{TS} & $76.17_{.05}$ & $2.09_{.3}$ & $2.02_{.23}$ & $75.63_{.11}$ & $1.85_{.1}$ & $1.83_{.07}$ & $84.35_{.06}$ & $1.35_{.08}$ & $1.35_{.08}$ & $85.59_{.05}$ & $1.33_{.18}$ & $1.32_{.19}$  \\ 
            \multirow{1}{*}{\textbf{+\textsc{Tna}(sparse)}} & $76.11_{.06}$ & $2.02_{.27}$ & $2.0_{.25}$ & $75.46_{.11} $& $1.78_{.05} $& $1.77_{.05} $& $84.34_{.07} $& $1.31_{.12} $& $1.31_{.08} $&  $85.5_{.03} $& $1.34_{.16} $& $1.28_{.18} $  \\
            \multirow{1}{*}{\textbf{+\textsc{Tna}(comp.)}} & $76.08_{.26}$ & $1.93_{.11}$ & $1.89_{.04}$ & $75.45_{.16}$ & $1.78_{.13}$ & $1.78_{.07}$ & $84.48_{.13}$ & $1.25_{.04}$ & $1.3_{.16}$ & $85.48_{.15}$ & $1.31_{.11}$ & $1.23_{.01}$ \\ \cmidrule(lr){1-13}
            
            \multirow{1}{*}{ETS} & $76.17_{.05}$ & $1.1_{.03}$ & $1.26_{.1}$ & $75.63_{.11}$ & $0.88_{.11}$ & $1.06_{.19}$ &  $84.35_{.06}$ & $0.95_{.06}$ & $1.14_{.06}$ &  $85.59_{.05}$ & $0.63_{.13}$ & $0.91_{.13}$  \\
            \multirow{1}{*}{\textbf{+\textsc{Tna}(sparse)}} & $76.11_{.06}$ & $1.06_{.05}$& $1.22_{.06}$&  $75.46_{.11}$& $0.79_{.13}$& $1.02_{.12}$&  $84.34_{.07}$& \textbf{0.9}$_{.05}$& \textbf{1.09}$_{.06}$&  $85.5_{.03}$& \textbf{0.6}$_{.14}$& $0.85_{.14}$  \\ 
            \multirow{1}{*}{\textbf{+\textsc{Tna}(comp.)}} & $76.04_{.11}$ & \textbf{1.01}$_{.15}$ & \textbf{1.19}$_{.03}$ & $75.45_{.23}$ & \textbf{0.78}$_{.11}$ & \textbf{1.01}$_{.14}$ & $84.5_{.11}$ & \textbf{0.9}$_{.17}$ & $1.1_{.11}$ & $85.43_{.12}$ & \textbf{0.6}$_{.05}$ & \textbf{0.84}$_{.07}$ \\ \cmidrule(lr){1-13} 
            
            \multirow{1}{*}{AAR} & $75.55_{.05}$ & $2.53_{.17}$ & $2.5_{.17}$  & $75.37_{.09}$ & $2.2_{.22}$ & $2.2_{.28}$ &$83.93_{.08}$ & $2.43_{.15}$ & $2.4_{.14}$ &  $84.96_{.13}$ & $2.36_{.28}$ & $2.2_{.37}$   \\
            \multirow{1}{*}{\textbf{+\textsc{Tna}(sparse)}} & $75.48_{.04}$ & $2.52_{.2}$ & $2.49_{.2}$ & $75.23_{.15}$& $2.33_{.2}$& $2.22_{.21} $&$83.93_{.07}$& $2.41_{.15}$& $2.38_{.14}$& $84.9_{.14}$& $2.3_{.34}$& $2.25_{.3}$\\
            \multirow{1}{*}{\textbf{+\textsc{Tna}(comp.)}} & $75.36_{.23}$ & $2.40_{.21}$ & $2.39_{.11}$ & $75.17_{.18}$ & $2.28_{.11}$ & $2.25_{.07}$ & $84.2_{.07}$& $2.38_{.04}$ & $2.38_{.11}$ & $84.81_{.12}$ & $2.32_{.03}$ & $2.28_{.05}$\\ 
            \bottomrule
            	\end{tabular}
            }
   
\end{table*}

\begin{figure*}
    \centering
    \includegraphics[width=0.85\linewidth]{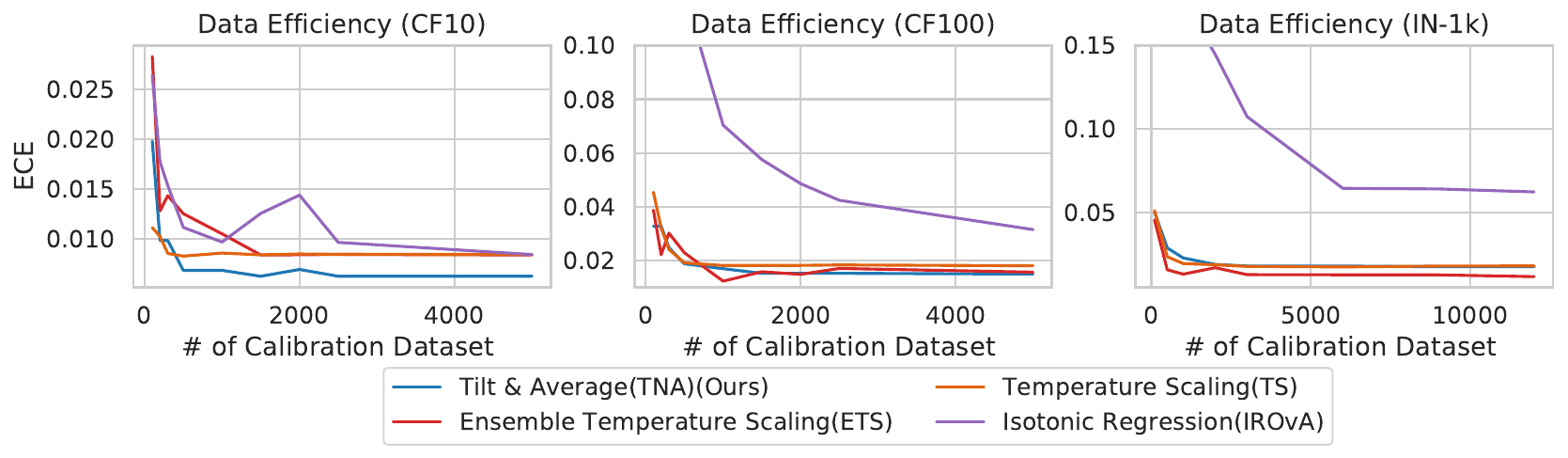}
    \caption{The data efficiency plot comparing 3 datasets(CIFAR10, CIFAR100, ImageNet). The architectures of WideResNet28x10, MobileNetv2, and ResNet50 are used respectively, when \textsc{Tna} is applied to the original weight. We demonstrate that \textsc{Tna} is efficient in data, requiring less additional calibration set.}
    \label{fig:data_efficiency}
\end{figure*}

\textbf{Models.} We compare a total of four model architectures for each dataset. For datasets SVHN and CIFARs, we take the models of WideResNet28x10\cite{zagoruyko2016wide}, MobileNetV2\cite{Sandler2018MobileNetV2}, PreResNet110\cite{he2016identity}, and GoogleNet\cite{szegedy2015going}. For ImageNet, we take 2 CNNs, ResNet-50\cite{he2016deep}, and 2 visual transformers ViT-L/16, ViT-H/14\cite{dosovitskiy2021an}. For the ImageNet trained networks, the models are borrowed from torchvision \citep{Marcel2010Torchvision} and timm\citep{rw2019timm}. For the other datasets, we follow the training details from \cite{ashukha2020pitfalls}, which are listed in the appendix.

\textbf{Methods.} For each dataset-model combination, we evaluate five baselines, including the uncalibrated model, and additionally assess nine methods with the proposed technique \textsc{Tna}, presented in this paper. The optimization is done by the calibration split, and the evaluation is done on the test split. The code will be available, \href{https://github.com/GYYYYYUUUUU/TNA_Angular_Scaling}{URL}.
\begin{itemize}
\item Baseline Method :  We apply the proposed method on the original weight, in combination with the calibration-map-based methods of IROvA\cite{Zadrozny2001Learning}, Temperature Scaling(TS)\cite{Guo17On}, Ensemble Temperature Scaling(ETS) \cite{zhang2020mix}, and Atypicality-Aware Recalibration(AAR) \cite{yuksekgonul2023beyond}.
\item \textsc{Tna}, \textsc{Tna}(Sparse) : 
\textsc{Tna}, \textsc{Tna}(Sparse) reveals outcomes achieved by optimizing each of the two steps independently: first, the alternative weight $\mathbf{W'}$, and subsequently, the calibration map $f_{cal}$. In the quest for the optimal alternative weight $\mathbf{W'}$, the initial step involves searching for the best angle $mRC$ within the range [0°, 90°] using the calibration set. As indicated in Proposition~\ref{prop: confidence_relaxation}, an increase in mRC results in relaxed confidence, rendering the ECE plot unimodal with mild assumptions into account. 
This can also be confirmed experimentally. Refer to Fig.~\ref{fig:Optimization_Curve} for detailed plots of the optimization curves on CIFAR10, CIFAR100, and ImageNet. Computationally, this process takes less than 5 minutes on a single GPU of RTX3080. (CIFAR10, and CIFAR100 take seconds.)

\item \textsc{Tna}(Comp.) : 
For the complete search, we aim to search for the best combination of the alternative weights and the selected calibration-map-based method. Using the calibration dataset, we perform a grid search for all possible $(\mathbf{W'}, f_{cal})$ pairs. Computationally, this process takes less than 30 minutes on a single GPU of RTX3080. (CIFAR10, CIFAR100 takes a few minutes.)

\end{itemize}

\textbf{Hyperparameters.} As proposed in the algorithm, we search over $\theta \in [0\degree, 90\degree]$. Over datasets and models, $n_e$ is set to 10, $\alpha=5$, $\beta=1$. The search interval of rotation number $n_t$ is set to 50. Unless noted, the maximum rotated angle $\theta_s$ is set to 0.9rad. The displayed value reflects an average of 5 repeated experiments, with the standard deviation depicted as a subscripted value. The number of bins $B$ is set to 15.

\subsection{Results}
The classifier weights newly derived through \textsc{Tna} do induce a very slight change in accuracy, albeit very marginal (less than 0.3\%). 
In terms of calibration performance, the application of weights by \textsc{Tna} surpasses the performance of using the original weight. Also in most cases, applying \textsc{Tna} in addition to the method improves performance than the case of applying the calibration-map-based method alone. The results indicate that, in addition to the conventional method of finding $f_{cal}$, it offers a broader solution space.

One interesting observation is that the combination of None+\textsc{Tna}, which only applies to adjust to the angle, outperforms Temperature Scaling. As highlighted in \cref{sec: discussion}, temperature scaling can be seen as adjusting for magnitude, and it is evident that adjusting for angles is particularly effective in cases where it matters.

Note that the TS and ETS do not affect the accuracy, while IROvA and AAR slightly have different values. This is because TS and ETS preserve the order of the class-wise element of the logit vector. On the contrary, the approaches of IROvA and AAR may exhibit a slight variation in accuracy, as the final probability is adjusted due to factors that can alter the order.

\subsection{Ablation Study}

\textbf{Data Efficiency.} Assessing data efficiency is also an important aspect \citep{zhang2020mix} to make the calibration process utilizable since recalibration requires an additional 'calibration set' other than a 'training set' to tune the probability. According to Fig.~\ref{fig:data_efficiency}, note that TS\cite{Guo17On} and the \textsc{Tna} show similar data efficiency to reach the optimal ECE while exhibiting better efficiency than IROvA\cite{Zadrozny2002Transforming}. One explanation for the observed phenomenon is that for \textsc{Tna}, we optimize a single parameter $\theta$(\textit{mRC}), similar to that of TS, in which we also optimize a single parameter $T$(temperature).

\textbf{Assumptions.} In our study, various assumptions were made to demonstrate that the algorithm has confidence calibration effects. We reiterate some of the assumptions and provide deeper explanations. In detail, we elucidate the assumptions of 1) $\angle(\mathbf{w}_i, R\mathbf{w}_i)$ is equal across classes $i\in[C]$, and 2) the feature space is high-dimensional with experimental results in the ~\cref{ablation: anal}.

\section{Discussion \& More Related Works}

\label{sec: discussion}

\textbf{Scaling Angles.} From a geometrical standpoint, we provide a technique that directly controls the angles. The role of the angle as a similarity/distance measure has been discussed in previous research \cite{Kansizoglou22Deep, Chen2020Angular, Kang20Decoupling, Peng22Angular}. However, to our knowledge, our paper is the first to introduce a geometric transformation that alters the angle between the class vector and the penultimate feature to adjust the confidence.

\textbf{Recalibration Approaches with Magnitude.}  The magnitude of the class vector also influences the confidence factor (Eq. \ref{eq: confidence}). Upon closer examination, this approach shares common ground with previously studied methods such as temperature scaling\cite{Guo17On}. As a straightforward proposition, temperature scaling applies scalar division on the logit vector $\mathbf{s}$ to obtain confidence. Substituting the $\textbf{w}_i$ as $\textbf{w}_i/T$, finding the optimal temperature has the effect of confidence relaxation by adjusting the magnitude. ETS\cite{zhang2020mix} and AAR\cite{yuksekgonul2023beyond} extend the concept of temperature scaling by seeking the optimal temperature sample-wise, considering the diversity of samples.

\textbf{Blessings of the dimensionality. } Our approach is founded upon a modicum of randomness as it pursues a rotation transformation.
Due to the \textit{blessings of the dimensionality} ($n$ of few hundred or thousands), we have a very low probability to rotate to the unwanted direction, where other class vectors might gain their score significantly and the overconfident vector loses it. From the high dimensional geometry \citep{blum_hopcroft_kannan_2020}, it is known that most of the volume is concentrated on the equator and the surface, and thus two arbitrary vectors are likely to be orthogonal. For example, in 2000-dimensional unit-ball, the volume near the equator of height $\frac{1}{2}$ is near $1-\frac{2}{10\sqrt{5}}e^{-\frac{1}{2}*{(10\sqrt{5}}^2)}\simeq(100-3.4*10^{-105})\%$ of the whole volume. So the single row vector $\mathbf{w}_i$ from the perspective of the $\mathbf{z}$ which has angle $\angle(\mathbf{w}_i, \mathbf{z})$ near $60 \degree$ with a perturbation of some angle is more likely to lean toward the equator, not the direction toward the $\mathbf{z}$. Note that ResNet-50, the dimension of the $\mathbf{z}$ space is 2048, $\cos{60}\degree=\frac{1}{2}$.

\section{Conclusion}
We introduce a fresh perspective on tackling the longstanding issue of recalibration by geometrically modifying the weights of the last layer instead of proposing a new calibration map for better calibration performance. We propose a new algorithm from this perspective, establish its theoretical background, and experimentally validate it. The presented technique outperforms existing methods.

\section{Broader Impact}
This paper presents work whose goal is to advance the field of Machine Learning. There are many potential societal consequences of our work, none which we feel must be specifically highlighted here.

\section*{Acknowledgements}

This work was partially supported by Korea Institute of Marine Science \& Technology Promotion(KIMST) funded by the Korea Coast Guard(RS-2023-00238652, Integrated Satellite-based Applications Development for Korea Coast Guard). Also, this work was partially supported by Korea Research Institute for defense Technology planning and advancement – Grant funded by Defense Acquisition Program Adminitration (DAPA) (KRIT-CT-23-020)

\section*{Impact Statement}

This paper presents work whose goal is to advance the field of 
Machine Learning. There are many potential societal consequences 
of our work, none which we feel must be specifically highlighted here.

\nocite{langley00}

\bibliography{main}

\begin{thebibliography}{44}
\providecommand{\natexlab}[1]{#1}
\providecommand{\url}[1]{\texttt{#1}}
\expandafter\ifx\csname urlstyle\endcsname\relax
  \providecommand{\doi}[1]{doi: #1}\else
  \providecommand{\doi}{doi: \begingroup \urlstyle{rm}\Url}\fi

\bibitem[Ashukha et~al.(2020)Ashukha, Lyzhov, Molchanov, and Vetrov]{ashukha2020pitfalls}
Ashukha, A., Lyzhov, A., Molchanov, D., and Vetrov, D.
\newblock Pitfalls of in-domain uncertainty estimation and ensembling in deep learning.
\newblock \emph{arXiv preprint arXiv:2002.06470}, 2020.

\bibitem[Blum et~al.(2020)Blum, Hopcroft, and Kannan]{blum_hopcroft_kannan_2020}
Blum, A., Hopcroft, J., and Kannan, R.
\newblock \emph{Foundations of Data Science}.
\newblock Cambridge University Press, 2020.
\newblock \doi{10.1017/9781108755528}.

\bibitem[Chen et~al.(2020)Chen, Liu, Yu, Kautz, Shrivastava, Garg, and Anandkumar]{Chen2020Angular}
Chen, B., Liu, W., Yu, Z., Kautz, J., Shrivastava, A., Garg, A., and Anandkumar, A.
\newblock Angular visual hardness.
\newblock In \emph{Proceedings of the 37th International Conference on Machine Learning, {ICML} 2020, 13-18 July 2020, Virtual Event}, volume 119 of \emph{Proceedings of Machine Learning Research}, pp.\  1637--1648. {PMLR}, 2020.
\newblock URL \url{http://proceedings.mlr.press/v119/chen20n.html}.

\bibitem[Daniel~Kermany(2018)]{Kermany2018OCT}
Daniel~Kermany, Kang~Zhang, M.~G.
\newblock Labeled optical coherence tomography (oct) and chest x-ray images for classification.
\newblock \emph{Mendeley Data}, 2018.

\bibitem[Deng et~al.(2009)Deng, Dong, Socher, Li, Li, and Fei-Fei]{deng2009imagenet}
Deng, J., Dong, W., Socher, R., Li, L.-J., Li, K., and Fei-Fei, L.
\newblock {ImageNet}: A large-scale hierarchical image database.
\newblock In \emph{CVPR}, 2009.

\bibitem[Ding et~al.(2021)Ding, Han, Liu, and Niethammer]{ding2021local}
Ding, Z., Han, X., Liu, P., and Niethammer, M.
\newblock Local temperature scaling for probability calibration.
\newblock In \emph{Proceedings of the IEEE/CVF International Conference on Computer Vision}, pp.\  6889--6899, 2021.

\bibitem[Dosovitskiy et~al.(2021)Dosovitskiy, Beyer, Kolesnikov, Weissenborn, Zhai, Unterthiner, Dehghani, Minderer, Heigold, Gelly, Uszkoreit, and Houlsby]{dosovitskiy2021an}
Dosovitskiy, A., Beyer, L., Kolesnikov, A., Weissenborn, D., Zhai, X., Unterthiner, T., Dehghani, M., Minderer, M., Heigold, G., Gelly, S., Uszkoreit, J., and Houlsby, N.
\newblock An image is worth 16x16 words: Transformers for image recognition at scale.
\newblock In \emph{ICLR}, 2021.

\bibitem[Euler(1776)]{EulerRotation}
Euler, L.
\newblock \emph{Novi commentarii Academiae Scientiarum Imperialis Petropolitanae}, volume~20.
\newblock Petropolis, Typis Academiae Scientarum, 1750-76, 1776.

\bibitem[Garipov et~al.(2018)Garipov, Izmailov, Podoprikhin, Vetrov, and Wilson]{Garipov2018Loss}
Garipov, T., Izmailov, P., Podoprikhin, D., Vetrov, D., and Wilson, A.~G.
\newblock Loss surfaces, mode connectivity, and fast ensembling of dnns.
\newblock In \emph{Proceedings of the 32nd International Conference on Neural Information Processing Systems}, NIPS'18, pp.\  8803–8812, Red Hook, NY, USA, 2018. Curran Associates Inc.

\bibitem[Givens(1958)]{Givens1958}
Givens, W.
\newblock Computation of plane unitary rotations transforming a general matrix to triangular form.
\newblock \emph{Journal of the Society for Industrial and Applied Mathematics}, 6\penalty0 (1):\penalty0 26--50, 1958.
\newblock ISSN 03684245.
\newblock URL \url{http://www.jstor.org/stable/2098861}.

\bibitem[Guo et~al.(2017)Guo, Pleiss, Sun, and Weinberger]{Guo17On}
Guo, C., Pleiss, G., Sun, Y., and Weinberger, K.~Q.
\newblock On calibration of modern neural networks.
\newblock In \emph{Proceedings of the 34th International Conference on Machine Learning - Volume 70}, ICML'17, pp.\  1321–1330. JMLR.org, 2017.

\bibitem[He et~al.(2016{\natexlab{a}})He, Zhang, Ren, and Sun]{he2016deep}
He, K., Zhang, X., Ren, S., and Sun, J.
\newblock Deep residual learning for image recognition.
\newblock In \emph{CVPR}, 2016{\natexlab{a}}.

\bibitem[He et~al.(2016{\natexlab{b}})He, Zhang, Ren, and Sun]{he2016identity}
He, K., Zhang, X., Ren, S., and Sun, J.
\newblock Identity mappings in deep residual networks, 2016{\natexlab{b}}.
\newblock URL \url{http://arxiv.org/abs/1603.05027}.
\newblock cite arxiv:1603.05027Comment: ECCV 2016 camera-ready.

\bibitem[Huang et~al.(2017)Huang, Liu, Van Der~Maaten, and Weinberger]{huang2017densely}
Huang, G., Liu, Z., Van Der~Maaten, L., and Weinberger, K.~Q.
\newblock Densely connected convolutional networks.
\newblock In \emph{CVPR}, 2017.

\bibitem[Hub(2022)]{EggData}
Hub, A.
\newblock {Mono and Colored Egg Data}.
\newblock \url{https://www.aihub.or.kr/aihubdata/data/view.do?currMenu=115&topMenu=100&dataSetSn=71504}, 2022.

\bibitem[Joy et~al.(2023)Joy, Pinto, Lim, Torr, and Dokania]{Joy2023sampleATS}
Joy, T., Pinto, F., Lim, S.-N., Torr, P. H.~S., and Dokania, P.~K.
\newblock Sample-dependent adaptive temperature scaling for improved calibration.
\newblock In \emph{Proceedings of the Thirty-Seventh AAAI Conference on Artificial Intelligence and Thirty-Fifth Conference on Innovative Applications of Artificial Intelligence and Thirteenth Symposium on Educational Advances in Artificial Intelligence}, AAAI'23/IAAI'23/EAAI'23. AAAI Press, 2023.
\newblock ISBN 978-1-57735-880-0.
\newblock \doi{10.1609/aaai.v37i12.26742}.
\newblock URL \url{https://doi.org/10.1609/aaai.v37i12.26742}.

\bibitem[Kang et~al.(2020)Kang, Xie, Rohrbach, Yan, Gordo, Feng, and Kalantidis]{Kang20Decoupling}
Kang, B., Xie, S., Rohrbach, M., Yan, Z., Gordo, A., Feng, J., and Kalantidis, Y.
\newblock Decoupling representation and classifier for long-tailed recognition.
\newblock In \emph{8th International Conference on Learning Representations, {ICLR} 2020, Addis Ababa, Ethiopia, April 26-30, 2020}. OpenReview.net, 2020.
\newblock URL \url{https://openreview.net/forum?id=r1gRTCVFvB}.

\bibitem[Kansizoglou et~al.(2022)Kansizoglou, Bampis, and Gasteratos]{Kansizoglou22Deep}
Kansizoglou, I., Bampis, L., and Gasteratos, A.
\newblock Deep feature space: A geometrical perspective.
\newblock \emph{IEEE Trans. Pattern Anal. Mach. Intell.}, 44\penalty0 (10Part2):\penalty0 6823–6838, oct 2022.
\newblock ISSN 0162-8828.
\newblock \doi{10.1109/TPAMI.2021.3094625}.
\newblock URL \url{https://doi.org/10.1109/TPAMI.2021.3094625}.

\bibitem[Krizhevsky et~al.(2010)Krizhevsky, Nair, and Hinton]{krizhevsky2010cifar}
Krizhevsky, A., Nair, V., and Hinton, G.
\newblock Cifar-10 (canadian institute for advanced research).
\newblock \emph{URL http://www. cs. toronto. edu/kriz/cifar. html}, 2010.

\bibitem[Kull et~al.(2019)Kull, Perello-Nieto, K\"{a}ngsepp, Filho, Song, and Flach]{Kull2019Beyond}
Kull, M., Perello-Nieto, M., K\"{a}ngsepp, M., Filho, T.~S., Song, H., and Flach, P.
\newblock \emph{Beyond Temperature Scaling: Obtaining Well-Calibrated Multiclass Probabilities with Dirichlet Calibration}.
\newblock Curran Associates Inc., Red Hook, NY, USA, 2019.

\bibitem[Li \& Vasconcelos(2019)Li and Vasconcelos]{li2019efficient}
Li, Y. and Vasconcelos, N.
\newblock Efficient multi-domain learning by covariance normalization.
\newblock In \emph{CVPR}, 2019.

\bibitem[Litjens et~al.(2017)Litjens, Kooi, Bejnordi, Setio, Ciompi, Ghafoorian, {van der Laak}, {van Ginneken}, and Sánchez]{Litjens2017Medical}
Litjens, G., Kooi, T., Bejnordi, B.~E., Setio, A. A.~A., Ciompi, F., Ghafoorian, M., {van der Laak}, J.~A., {van Ginneken}, B., and Sánchez, C.~I.
\newblock A survey on deep learning in medical image analysis.
\newblock \emph{Medical Image Analysis}, 42:\penalty0 60--88, 2017.
\newblock ISSN 1361-8415.
\newblock \doi{https://doi.org/10.1016/j.media.2017.07.005}.
\newblock URL \url{https://www.sciencedirect.com/science/article/pii/S1361841517301135}.

\bibitem[Ma \& Blaschko(2021)Ma and Blaschko]{Ma2021a}
Ma, X. and Blaschko, M.~B.
\newblock Meta-cal: Well-controlled post-hoc calibration by ranking.
\newblock In \emph{International Conference on Machine Learning}, 2021.

\bibitem[Marcel \& Rodriguez(2010)Marcel and Rodriguez]{Marcel2010Torchvision}
Marcel, S. and Rodriguez, Y.
\newblock Torchvision the machine-vision package of torch.
\newblock pp.\  1485--1488, 10 2010.
\newblock \doi{10.1145/1873951.1874254}.

\bibitem[Moskola{\"\i} et~al.(2021)Moskola{\"\i}, Abdou, and Dipanda]{moskolai2021satellite}
Moskola{\"\i}, W.~R., Abdou, W., and Dipanda, A.
\newblock Application of deep learning architectures for satellite image time series prediction: A review.
\newblock \emph{Remote Sensing}, 13\penalty0 (23):\penalty0 4822, 2021.

\bibitem[Mukhoti et~al.(2020)Mukhoti, Kulharia, Sanyal, Golodetz, Torr, and Dokania]{mukhoti2020calibrating}
Mukhoti, J., Kulharia, V., Sanyal, A., Golodetz, S., Torr, P.~H., and Dokania, P.~K.
\newblock Calibrating deep neural networks using focal loss.
\newblock 2020.

\bibitem[Naeini et~al.(2015)Naeini, Cooper, and Hauskrecht]{Naeini2015Obtaining}
Naeini, M., Cooper, G., and Hauskrecht, M.
\newblock Obtaining well calibrated probabilities using bayesian binning.
\newblock \emph{Proceedings of the AAAI Conference on Artificial Intelligence}, 29\penalty0 (1), Feb. 2015.
\newblock \doi{10.1609/aaai.v29i1.9602}.
\newblock URL \url{https://ojs.aaai.org/index.php/AAAI/article/view/9602}.

\bibitem[Netzer et~al.(2011)Netzer, Wang, Coates, Bissacco, Wu, and Ng]{netzer2011reading}
Netzer, Y., Wang, T., Coates, A., Bissacco, A., Wu, B., and Ng, A.~Y.
\newblock Reading digits in natural images with unsupervised feature learning.
\newblock In \emph{NeurIPS}, 2011.

\bibitem[Ovadia et~al.(2019)Ovadia, Fertig, Ren, Nado, Sculley, Nowozin, Dillon, Lakshminarayanan, and Snoek]{Ovadia2019Can}
Ovadia, Y., Fertig, E., Ren, J., Nado, Z., Sculley, D., Nowozin, S., Dillon, J.~V., Lakshminarayanan, B., and Snoek, J.
\newblock \emph{Can You Trust Your Model's Uncertainty? Evaluating Predictive Uncertainty under Dataset Shift}.
\newblock Curran Associates Inc., Red Hook, NY, USA, 2019.

\bibitem[Patel et~al.(2021)Patel, Beluch, Yang, Pfeiffer, and Zhang]{Patel2021Multi}
Patel, Beluch, W., Yang, B., Pfeiffer, M., and Zhang, D.
\newblock Multi-class uncertainty calibration via mutual information maximization-based binning.
\newblock In \emph{International Conference on Learning Representations(ICLR)}, 2021.

\bibitem[Peng et~al.(2022)Peng, Islam, and Tu]{Peng22Angular}
Peng, B., Islam, M., and Tu, M.
\newblock Angular gap: Reducing the uncertainty of image difficulty through model calibration.
\newblock In \emph{Proceedings of the 30th ACM International Conference on Multimedia}, MM '22, pp.\  979–987, New York, NY, USA, 2022. Association for Computing Machinery.
\newblock ISBN 9781450392037.
\newblock \doi{10.1145/3503161.3548289}.
\newblock URL \url{https://doi.org/10.1145/3503161.3548289}.

\bibitem[Platt(1999)]{Platt1999}
Platt, J.
\newblock Probabilities for sv machines.
\newblock pp.\  61--74, 01 1999.

\bibitem[Rahimi et~al.(2020)Rahimi, Shaban, Cheng, Hartley, and Boots]{Rahimi2020Intra}
Rahimi, A., Shaban, A., Cheng, C.-A., Hartley, R., and Boots, B.
\newblock Intra order-preserving functions for calibration of multi-class neural networks.
\newblock In \emph{Proceedings of the 34th International Conference on Neural Information Processing Systems}, NeurIPS'20, Red Hook, NY, USA, 2020. Curran Associates Inc.
\newblock ISBN 9781713829546.

\bibitem[Sandler et~al.(2018)Sandler, Howard, Zhu, Zhmoginov, and Chen]{Sandler2018MobileNetV2}
Sandler, M., Howard, A.~G., Zhu, M., Zhmoginov, A., and Chen, L.
\newblock Mobilenetv2: Inverted residuals and linear bottlenecks.
\newblock In \emph{2018 {IEEE} Conference on Computer Vision and Pattern Recognition, {CVPR} 2018, Salt Lake City, UT, USA, June 18-22, 2018}, pp.\  4510--4520. Computer Vision Foundation / {IEEE} Computer Society, 2018.
\newblock \doi{10.1109/CVPR.2018.00474}.
\newblock URL \url{http://openaccess.thecvf.com/content\_cvpr\_2018/html/Sandler\_MobileNetV2\_Inverted\_Residuals\_CVPR\_2018\_paper.html}.

\bibitem[Szegedy et~al.(2015)Szegedy, Liu, Jia, Sermanet, Reed, Anguelov, Erhan, Vanhoucke, and Rabinovich]{szegedy2015going}
Szegedy, C., Liu, W., Jia, Y., Sermanet, P., Reed, S., Anguelov, D., Erhan, D., Vanhoucke, V., and Rabinovich, A.
\newblock Going deeper with convolutions.
\newblock In \emph{CVPR}, 2015.

\bibitem[Tomani et~al.(2022)Tomani, Cremers, and Buettner]{Tomani2022ParameterizedTS}
Tomani, C., Cremers, D., and Buettner, F.
\newblock Parameterized temperature scaling for boosting the expressive power in post-hoc uncertainty calibration.
\newblock In \emph{European Conference on Computer Vision}, 2022.

\bibitem[Wenger et~al.(2020)Wenger, Kjellström, and Triebel]{Wenger2020Nonparametric}
Wenger, J., Kjellström, H., and Triebel, R.
\newblock Non-parametric calibration for classification.
\newblock AISTATs'20, 08 2020.

\bibitem[Wightman(2019)]{rw2019timm}
Wightman, R.
\newblock Pytorch image models.
\newblock \url{https://github.com/rwightman/pytorch-image-models}, 2019.

\bibitem[Yuksekgonul et~al.(2023)Yuksekgonul, Zhang, Zou, and Guestrin]{yuksekgonul2023beyond}
Yuksekgonul, M., Zhang, L., Zou, J., and Guestrin, C.
\newblock Beyond confidence: Reliable models should also consider atypicality.
\newblock \emph{arXiv preprint arXiv:2305.18262}, 2023.

\bibitem[Yurtsever et~al.(2020)Yurtsever, Lambert, Carballo, and Takeda]{Yurtsever2020Driving}
Yurtsever, E., Lambert, J., Carballo, A., and Takeda, K.
\newblock A survey of autonomous driving: Common practices and emerging technologies.
\newblock \emph{IEEE Access}, 8:\penalty0 58443--58469, 2020.
\newblock \doi{10.1109/ACCESS.2020.2983149}.

\bibitem[Zadrozny \& Elkan(2001)Zadrozny and Elkan]{Zadrozny2001Learning}
Zadrozny, B. and Elkan, C.
\newblock Learning and making decisions when costs and probabilities are both unknown.
\newblock In \emph{Proceedings of the Seventh ACM SIGKDD International Conference on Knowledge Discovery and Data Mining}, KDD '01, pp.\  204–213, New York, NY, USA, 2001. Association for Computing Machinery.
\newblock ISBN 158113391X.
\newblock \doi{10.1145/502512.502540}.
\newblock URL \url{https://doi.org/10.1145/502512.502540}.

\bibitem[Zadrozny \& Elkan(2002)Zadrozny and Elkan]{Zadrozny2002Transforming}
Zadrozny, B. and Elkan, C.
\newblock Transforming classifier scores into accurate multiclass probability estimates.
\newblock In \emph{Proceedings of the Eighth ACM SIGKDD International Conference on Knowledge Discovery and Data Mining}, KDD '02, pp.\  694–699, New York, NY, USA, 2002. Association for Computing Machinery.
\newblock ISBN 158113567X.
\newblock \doi{10.1145/775047.775151}.
\newblock URL \url{https://doi.org/10.1145/775047.775151}.

\bibitem[Zagoruyko \& Komodakis(2016)Zagoruyko and Komodakis]{zagoruyko2016wide}
Zagoruyko, S. and Komodakis, N.
\newblock Wide residual networks.
\newblock In \emph{BMVC}, 2016.

\bibitem[Zhang et~al.(2020)Zhang, Kailkhura, and Han]{zhang2020mix}
Zhang, J., Kailkhura, B., and Han, T.
\newblock Mix-n-match: Ensemble and compositional methods for uncertainty calibration in deep learning.
\newblock In \emph{International Conference on Machine Learning (ICML)}, 2020.

\end{thebibliography}
\bibliographystyle{icml2024}

\newpage
\appendix
\onecolumn

\section{Extended Experiments}
\subsection{(Extended Experiment on Tab.~\ref{tab:hardness}) Main Results on SVHN dataset.}

The table reports the results for the SVHN dataset in Tab.~\ref{tab:SVHN results}. Similar trends to those shown in Tab.~\ref{tab:hardness} are observed in this dataset as well.
\begin{table}
\caption{\textbf{(Extended Experiment on Tab.~\ref{tab:hardness})} Main results displaying changes after \textsc{Tna}, in accuracy and calibration performance. The experiments were conducted SVHN dataset expanded on the main result in Tab. ~\ref{tab:hardness}, evaluating four different model architectures for each dataset. In addition to applying five recalibration techniques, including the uncalibrated model as a baseline, we evaluate \textsc{Tna}(sparse) and \textsc{Tna}(comp.), which involve two stages of optimization and explore the entire search space, respectively. 
In a specific dataset-model combination, indicate the technique that achieved the best results among all methods in \textbf{bold}, and \underline{underline} the technique that achieved the best results without using proposed method. The results are displayed as averages over five runs.}\label{tab:SVHN results}
    \centering 
    \scalebox{0.65}{
    	\begin{tabular}{c|ccc|ccc|ccc|ccc}
    		\toprule
       \multicolumn{1}{c}{} & \multicolumn{3}{c}{\textbf{SVHN-WideResNet28x10}} & \multicolumn{3}{c}{\textbf{SVHN-MobileNetV2}} & \multicolumn{3}{c}{\textbf{SVHN-PreResNet110}}  & \multicolumn{3}{c}{\textbf{SVHN-GoogleNet}}  \\
                \midrule
    		\textbf{Methods} & \textbf{Acc.} ($\uparrow$) &  \textbf{ECE} ($\downarrow$) & \textbf{AdaECE} ($\downarrow$) & \textbf{Acc.} ($\uparrow$) &  \textbf{ECE} ($\downarrow$) & \textbf{AdaECE} ($\downarrow$) & \textbf{Acc.} ($\uparrow$) &  \textbf{ECE} ($\downarrow$) & \textbf{AdaECE} ($\downarrow$) & \textbf{Acc.} ($\uparrow$) &  \textbf{ECE} ($\downarrow$) & \textbf{AdaECE} ($\downarrow$) \\\cmidrule(lr){1-13}
    	\multirow{1}{*}{None} &$97.02_{.06}$ & $1.86_{.04}$ & $1.82_{.04}$ & $95.74_{.03}$ & $0.54_{.02}$ & $0.52_{.02}$ & $96.61_{.03}$ & $1.58_{.03}$ & $1.58_{.03}$ & $96.3_{.06}$ & $2.31_{.04}$ & $2.3_{.04}$ \\ 
            \multirow{1}{*}{\textbf{+\textsc{Tna}}} & $97.0_{.06} $& $0.45_{.01} $& $0.49_{.02} $&  $95.75_{.03} $& $0.43_{.02} $& $0.45_{.03} $& $96.56_{.03} $& $0.53_{.07} $& $0.74_{.04} $& $96.29_{.06} $& $0.59_{.03} $& $0.99_{.04} $

  \\ \cmidrule(lr){1-13}
            
             \multirow{1}{*}{IROvA} &$96.97_{.06}$ & $0.65_{.18}$ & $0.44_{.03}$ & $95.77_{.06}$ & $0.87_{.17}$ & $0.5_{.14}$ & $96.64_{.08}$ & $0.76_{.15}$ & $0.53_{.12}$ & $96.25_{.1}$ & $0.66_{.18}$ & $0.74_{.1}$ 
  \\
            \multirow{1}{*}{\textbf{+\textsc{Tna}(sparse)}} & $96.96_{.07} $& $0.55_{.22} $& $0.42_{.06} $&  $95.77_{.06} $& $0.84_{.19} $& $0.46_{.1} $& $96.59_{.08} $& $0.72_{.16} $& $0.51_{.14} $& $96.26_{.08} $& $0.64_{.18} $& $0.67_{.11} $
 \\ 
            \multirow{1}{*}{\textbf{+\textsc{Tna}(comp.)}} &  $96.99_{.04}$ & $0.49_{.15}$ & $0.42_{.11}$ & $95.81_{.05}$ & $0.77_{.17}$ & $0.44_{.1}$ & $96.58_{.06}$ & $0.71_{.22}$ & $0.54_{.09}$ & $96.28_{.03}$ & $0.55_{.01}$ & $0.63_{.04}$ 
 \\ \cmidrule(lr){1-13}
            
            \multirow{1}{*}{TS} & $97.02_{.06}$ & $0.67_{.08}$ & $0.55_{.04}$ & $95.74_{.03}$ & $0.33_{.05}$ & $0.28_{.06}$ & $96.61_{.03}$ & $0.76_{.19}$ & $0.87_{.12}$ & $96.3_{.06}$ & $1.01_{.14}$ & $1.07_{.06}$ 
\\ 
            \multirow{1}{*}{\textbf{+\textsc{Tna}(sparse)}} & $97.0_{.06} $& $0.61_{.1} $& $0.5_{.04} $&  $95.75_{.03} $& $0.32_{.05} $& $0.25_{.04} $& $96.56_{.03} $& $0.64_{.15} $& $0.83_{.11} $& $96.29_{.06} $& $0.97_{.12} $& $1.01_{.05} $ 
  \\ 
            \multirow{1}{*}{\textbf{+\textsc{Tna}(comp.)}} & $97.01_{.03}$ & $0.44_{.07}$ & $0.44_{.06}$ &  $95.75_{.03} $& $0.32_{.05} $& $0.25_{.04} $& $96.5_{.03}$ & $0.53_{.12}$ & $0.68_{.1}$ & $96.29_{.06} $& $0.97_{.12} $& $1.01_{.05} $ 

 \\ \cmidrule(lr){1-13}
            
            \multirow{1}{*}{ETS} & $97.02_{.06}$ & $0.67_{.08}$ & $0.55_{.04}$ & $95.74_{.03}$ & $0.34_{.05}$ & $0.28_{.05}$ &  $96.61_{.03}$ & $0.76_{.19}$ & $0.87_{.12}$ & $96.3_{.06}$ & $1.01_{.14}$ & $1.07_{.06}$ 
\\
            \multirow{1}{*}{\textbf{+\textsc{Tna}(sparse)}} & $97.0_{.06} $& $0.61_{.1} $& $0.5_{.04} $& $95.75_{.03} $& $0.33_{.05} $& $0.25_{.04} $&  $96.56_{.03} $& $0.65_{.14} $& $0.84_{.11} $& $96.29_{.06} $& $0.97_{.12} $& $1.01_{.05} $
\\ 
            \multirow{1}{*}{\textbf{+\textsc{Tna}(comp.)}} & $97.01_{.03}$ & $0.44_{.07}$ & $0.44_{.06}$ & $95.79_{.05}$ & $0.32_{.07}$ & $0.25_{.01}$ & $96.49_{.05}$ & $0.52_{.12}$ & $0.68_{.1}$ & $96.29_{.06} $& $0.97_{.12} $& $1.01_{.05} $

 \\ \cmidrule(lr){1-13}
            
            \multirow{1}{*}{AAR} & $97.04_{.05}$ & $0.32_{.12}$ & $0.29_{.11}$ &  $95.83_{.03}$ & $0.35_{.06}$ & $0.3_{.06}$ &  $96.66_{.06}$ & $0.51_{.1}$ & $0.48_{.16}$ & $96.32_{.07}$ & $0.43_{.08}$ & $0.43_{.09}$ 
 \\
            \multirow{1}{*}{\textbf{+\textsc{Tna}(sparse)}} & $97.03_{.06} $& $0.3_{.13} $& \textbf{0.28}$_{.11} $& $95.84_{.03} $& $0.34_{.06} $& $0.3_{.06} $& $96.65_{.06} $& $0.49_{.11} $& $0.47_{.13} $&  $96.31_{.06} $& \textbf{0.4}$_{.1} $& \textbf{0.42}$_{.09} $\\ 
            \multirow{1}{*}{\textbf{+\textsc{Tna}(comp.)}} & $97.0_{.04}$ & \textbf{0.3}$_{.1}$ & $0.29_{.08}$ & $95.88_{.04}$ & \textbf{0.27}$_{.1}$ & \textbf{0.29}$_{.07}$ & $96.62_{.06}$ & \textbf{0.41}$_{.11}$ & \textbf{0.38}$_{.05}$ &  $96.31_{.06} $& \textbf{0.4}$_{.1} $& \textbf{0.42}$_{.09} $

   \\
 
            \bottomrule
            	\end{tabular}
            }
    
\end{table}

\subsection{(Extended Experiment on Fig.~\ref{fig:accuracy_compensation}) Accuracy Compensation as increase of $n_e$ across datasets and models.}

\begin{figure}[ht]
\begin{center}
\includegraphics[width=0.4\columnwidth]{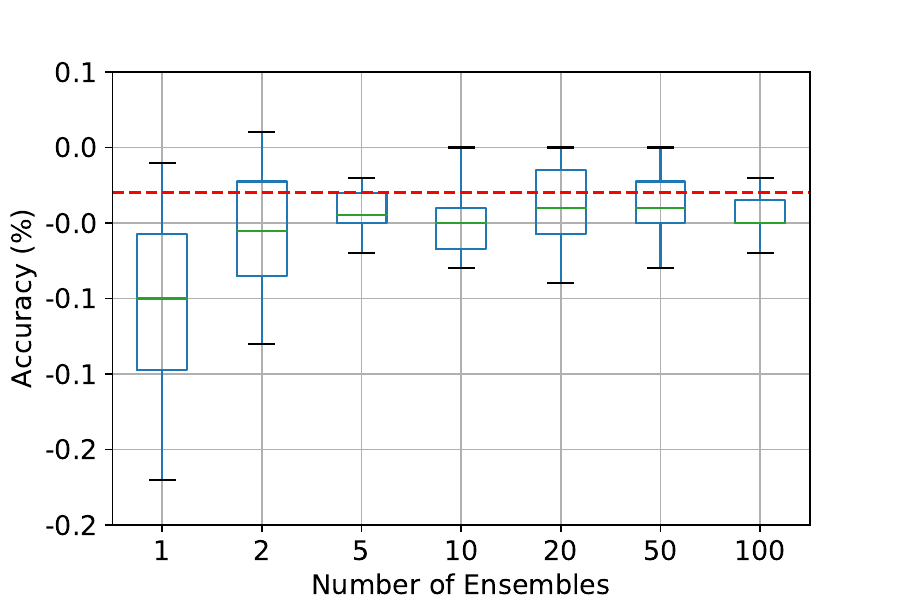}
\includegraphics[width=0.4\columnwidth]{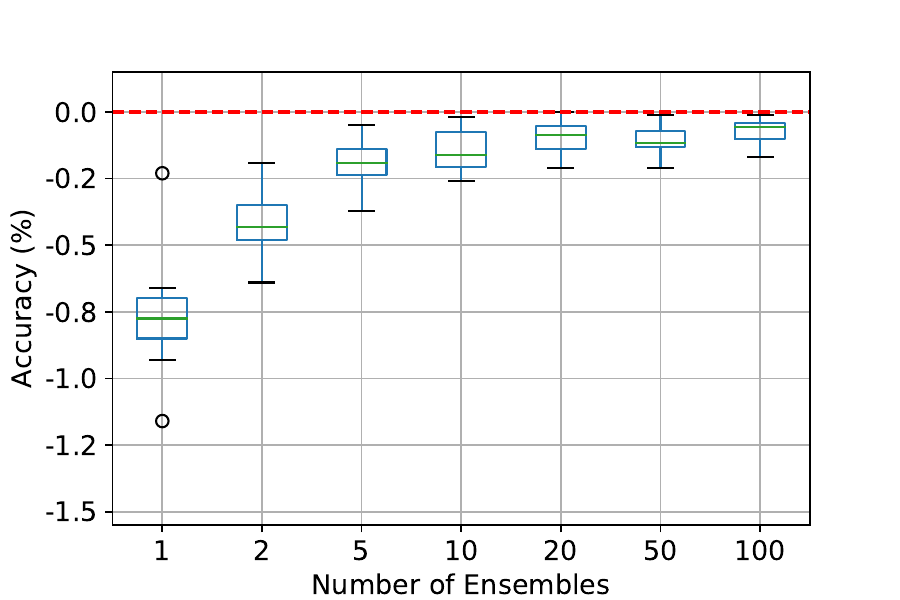}
\includegraphics[width=0.4\columnwidth]{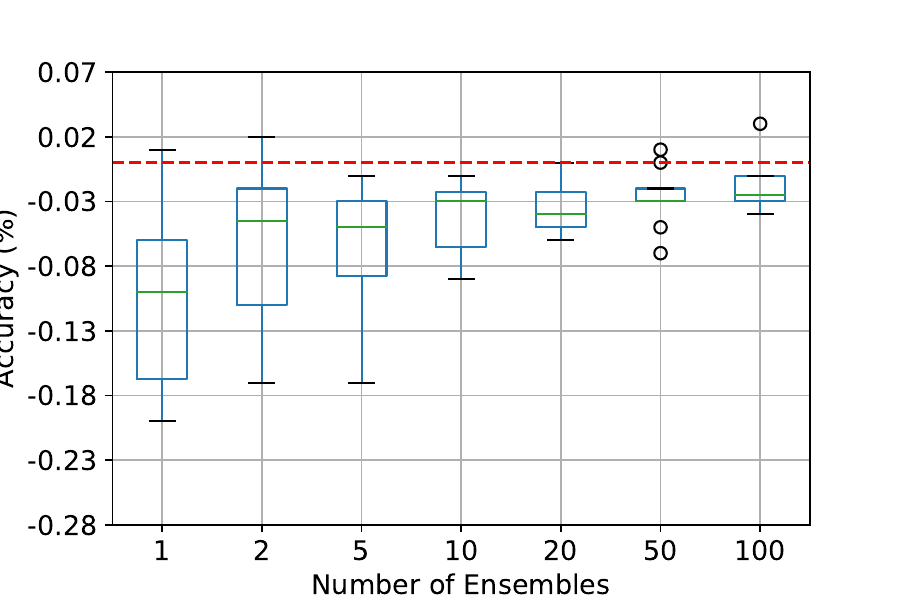}
\includegraphics[width=0.4\columnwidth]{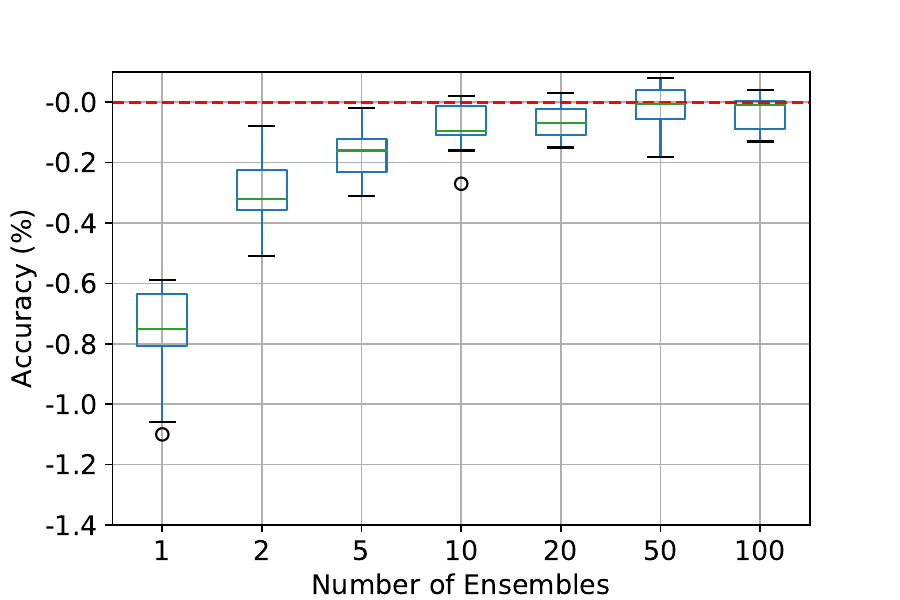}
\includegraphics[width=0.4\columnwidth]{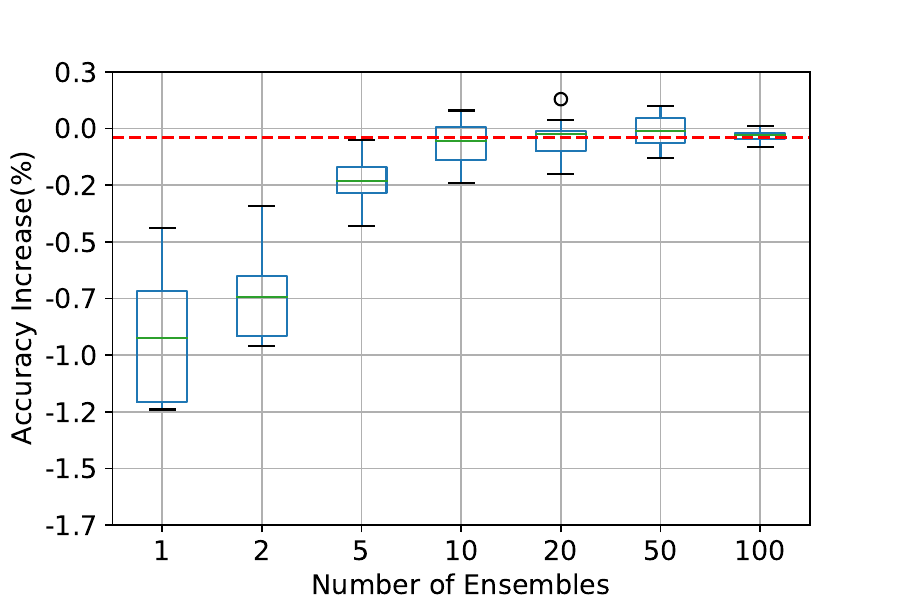}
\includegraphics[width=0.4\columnwidth]{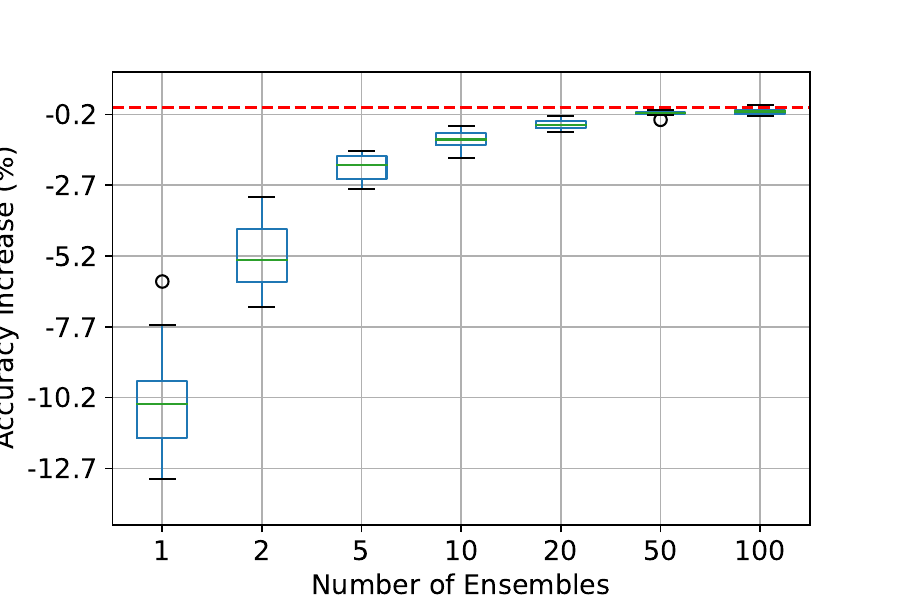}
\caption{\textbf{Extended experiment on Fig.~\ref{fig:accuracy_compensation}} on CIFARs. Plots of the accuracy of the ensembled outputs correspond to the number of ensemble members, on CIFAR10(left column) and CIFAR100(right column) dataset. The first row states the results on WideResNet28x10, middle row states the results on MobileNetV2, and the third row states the results on PreResNet110. The accuracy is well compensated as the $n_e$ increases.}
\label{fig:accuracy_compensation_supp_CF}
\end{center}
\end{figure}

In this section, to illustrate the ability to interpolate accuracy across different models, we conduct additional accuracy compensation experiments for two datasets, CIFAR10 (left column) and CIFAR100 (right column), considering three model architectures: WideResNet28x10 (upper), MobileNetV2 (middle), and PreResNet110 (lower). We increase the number $n_e$ and observe the interpolation of accuracy. To assess how accuracy is interpolated on average, we utilize boxplots and statistically plot accuracy interpolation by repeating the presentation algorithm 10 times. The red dashed line represents the performance of the original weights, and the numerical values on the left indicate the quantification of accuracy changes. Negative values indicate a degradation in accuracy, and as observed in the Fig.~\ref{fig:accuracy_compensation_supp_CF}, increasing $n_e$ leads to better accuracy interpolation.

\subsection{(Extended Experiment on Fig.~\ref{fig:data_efficiency}) The shifting effect of angles across dataset and models.}

\begin{figure}[ht]
\begin{center}
\centerline{\includegraphics[width=0.9\columnwidth]{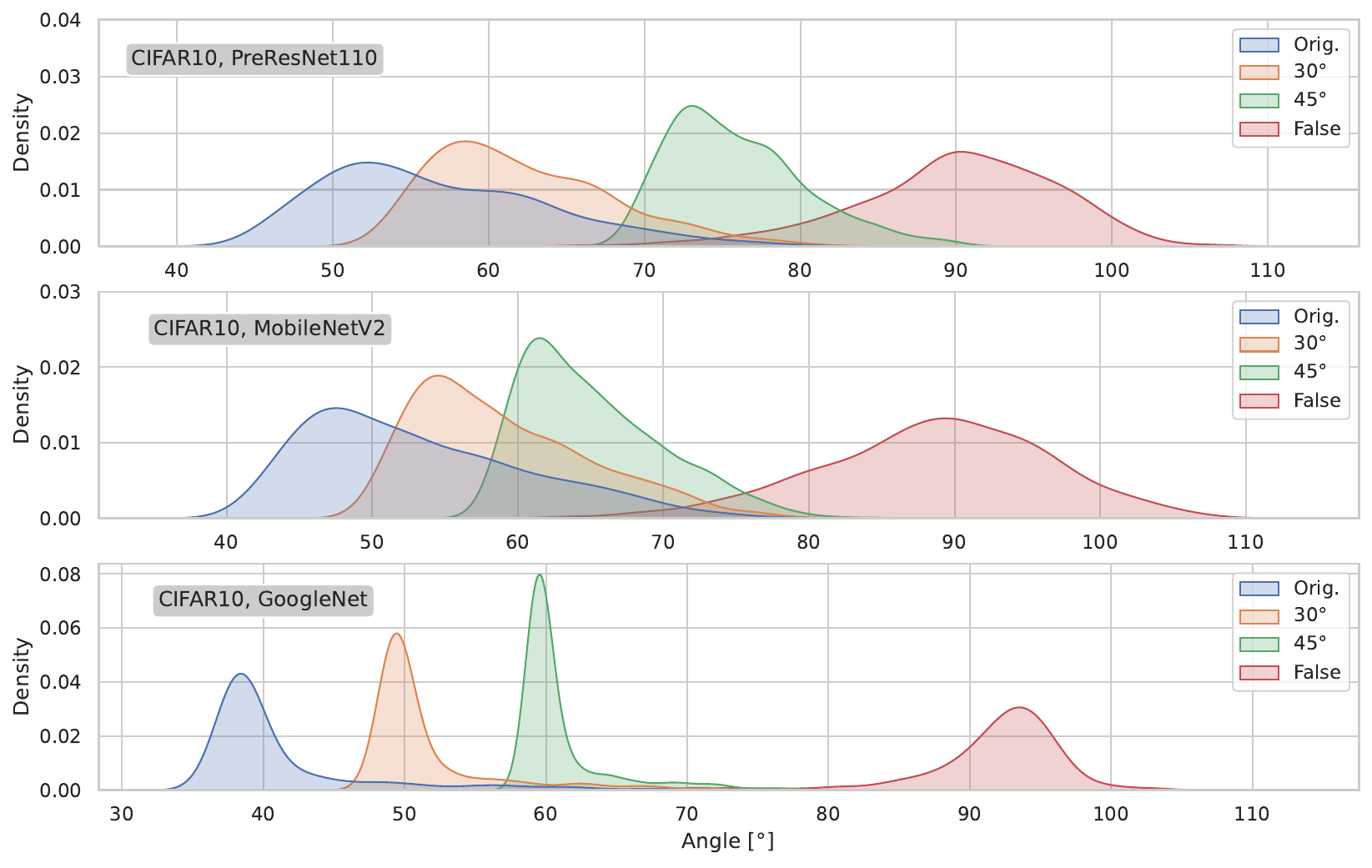}}
\centerline{\includegraphics[width=0.9\columnwidth]{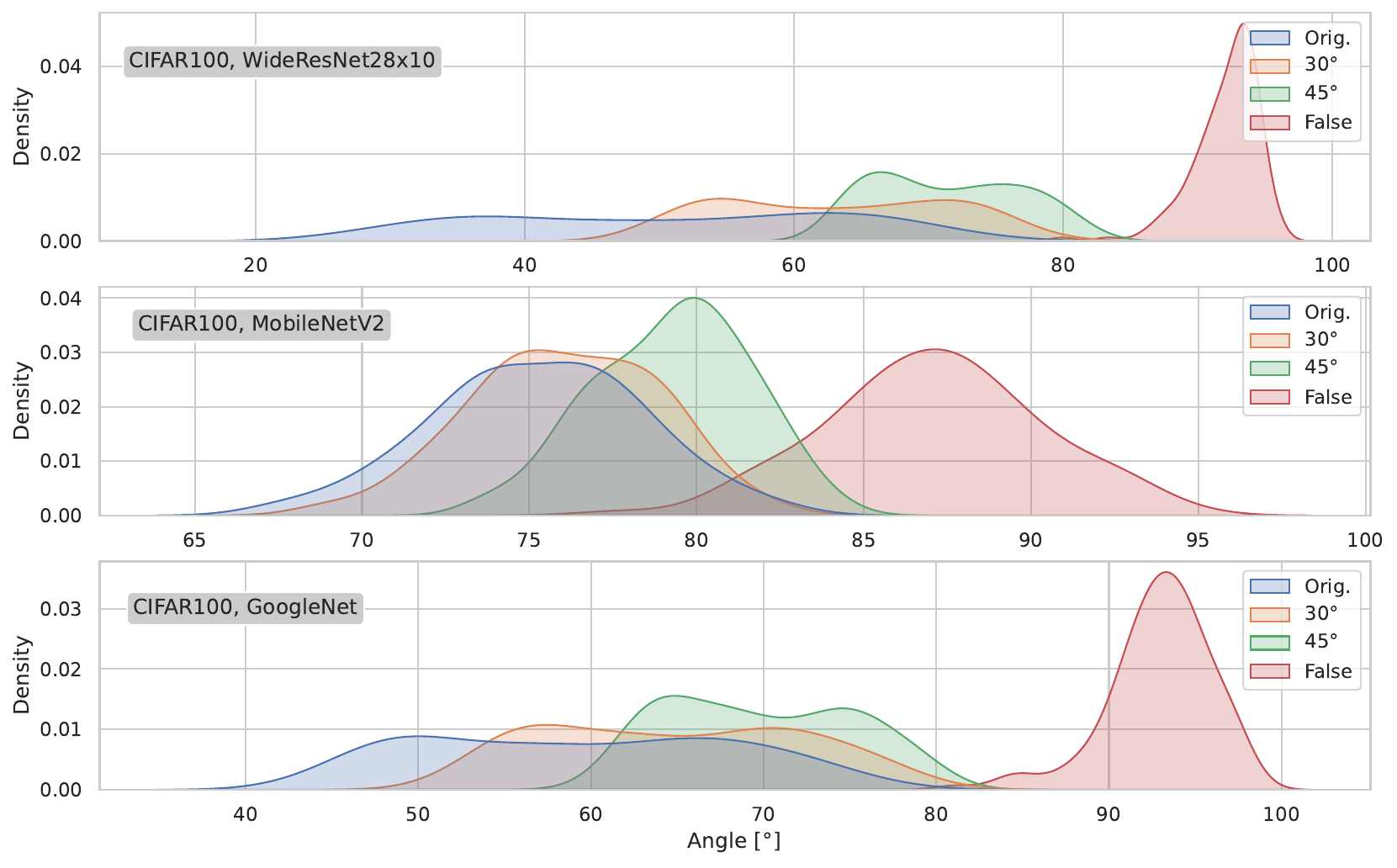}}
\caption{\textbf{Extended experiment on Fig.~\ref{fig:data_distribution}} on CIFARs. Distribution plot of the data samples with angle between class vector and pf, $\angle(\mathbf{w}_i, \mathbf{z}_x)$, with the corresponding predicted class row vector of the \textbf{original weight} (Orig.) and the \textbf{tilted weights} by depicted angle of $mRC$ (30\textdegree, 45\textdegree). ``False" denotes the angle of $pf$ with the class vector does not correspond to the respective class. As the $mRC$ increases, the angles shift towards 90 \textdegree. Best seen with colors.}
\label{fig:data_distribution_supp_CF}
\end{center}
\end{figure}

To demonstrate that the proposed angle-shifting effect exists across all datasets and models, additional figures are presented as follows. Firstly, in Fig.~\ref{fig:data_distribution_supp_CF}, plots for combinations not included in the main paper are provided for wideResNet28x10, MobileNetV2, PreResNet110, and GoogleNet on CIFAR10 and CIFAR100 datasets. The plots showcase the shifting effect of angles across models and datasets.

\begin{figure}[ht]
\begin{center}
\centerline{\includegraphics[width=0.9\columnwidth]{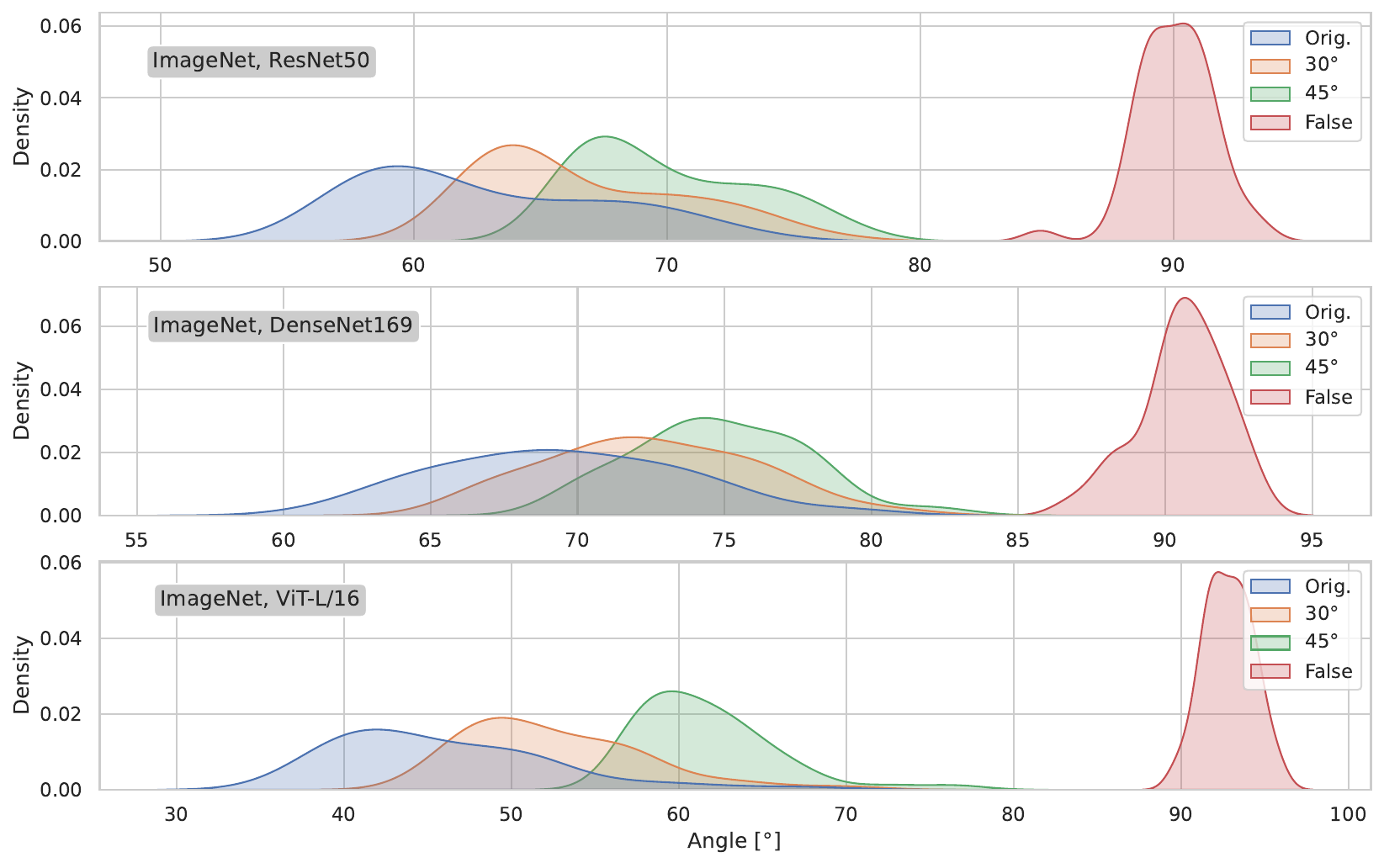}}
\caption{\textbf{Extended experiment on Fig.~\ref{fig:data_distribution}} on ImageNet dataset. Distribution plot of the data samples with angle between class vector and pf, $\angle(\mathbf{w}_i, \mathbf{z}_x)$, with the corresponding predicted class row vector of the \textbf{original weight} (Orig.) and the \textbf{tilted weights} by depicted angle of $mRC$ (30\textdegree, 45\textdegree). ``False" denotes the angle of $pf$ with the class vector does not correspond to the respective class. As the $mRC$ increases, the angles shift towards 90 \textdegree. Best seen with colors.}
\label{fig:data_distribution_supp_IN}
\end{center}
\end{figure}

\section{Further Analysis, Ablation Studies}

\label{ablation: anal}

\begin{figure}[h]
    \centering
    \includegraphics[width=1.0\linewidth]{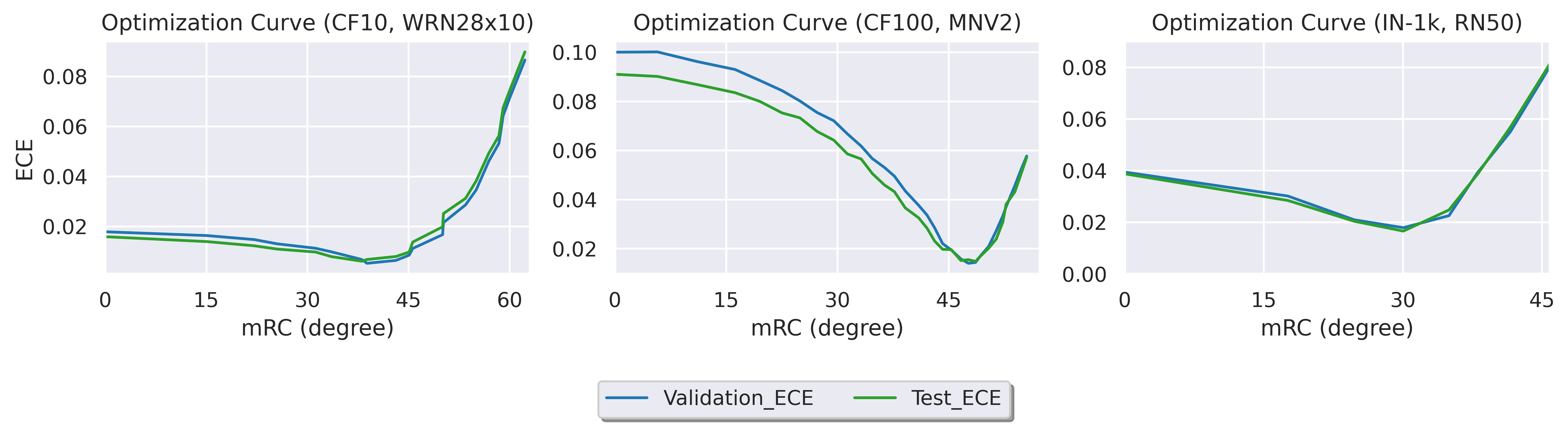}
    \caption{The optimization curve when \textsc{TNA} is done with number of ensembles $n_e=10$.  We leverage the calibration set to optimize and find the optimal trained weights, applying grid search with the calibration dataset's information allowed (`Validation\textunderscore ECE'). The `Test\textunderscore ECE', ECE on the test dataset, exhibits similar trend with the `Validation\textunderscore ECE'. Note that the curve is unimodal, making it able to optimize.}
    \label{fig:Optimization_Curve}
\end{figure}

\textbf{\textsc{Tna} : The optimization curve.} We report the results of \textsc{TNA} after weight averaging is done when using 10 ensembles, $n_e=10$ in Fig.~\ref{fig:Optimization_Curve}. We utilize the calibration set to optimize the process and determine the best trained weights. It is important to note that the curve is unimodal a, and thus it is quite straight-forward to find the optimal angle of $mRC$, when \textsc{Tna} is applied.

\textbf{Assumptions.} In our study, various assumptions were made to demonstrate that the algorithm has confidence calibration effects. We reiterate some of the assumptions and provide deeper explanations. In detail, we elucidate the assumptions of 1) $\angle(\mathbf{w}_i, R\mathbf{w}_i)$ is equal across classes $i\in[C]$, and 2) the dimension $n$ is high-dimensional in the appendix, ~\cref{ablation: anal}.

1) $\angle(\mathbf{w}_i, R\mathbf{w}_i)$ is equal across classes $i\in[C]$ : $\angle(\mathbf{w}_i, R\mathbf{w}_i)$ might vary across classes. This situation arises when the rotation transform generated from the proposed algorithm \textsc{Tilt} adjusts angles differently for each class. This violates the assumption in Prop.~\ref{prop: confidence_relaxation} which assumes $\forall i \in [C], \angle(\mathbf{w}_i, \mathbf{z})=\theta$. 
Experimentally, as shown in Fig.~\ref{fig:plot_compund_rotations}, the left column describes the mean of rotation over classes, while the right column describes the standard deviation of rotation over classes. One observation is that in all cases, the standard deviation is not large, but it can occur when the number of classes $C$ is small (i.e. CIFAR10, $C=10$), number of rotations $n_r$ is small, or $\theta_s$ is small. To address this issue and create a rotation corresponding to a specific $mRC(W,R)=\theta$, using a small value for $\theta_s$ and increasing the value of $n_r$ can be effective.

\begin{figure}
     \centering
     \includegraphics[width=0.9\linewidth]{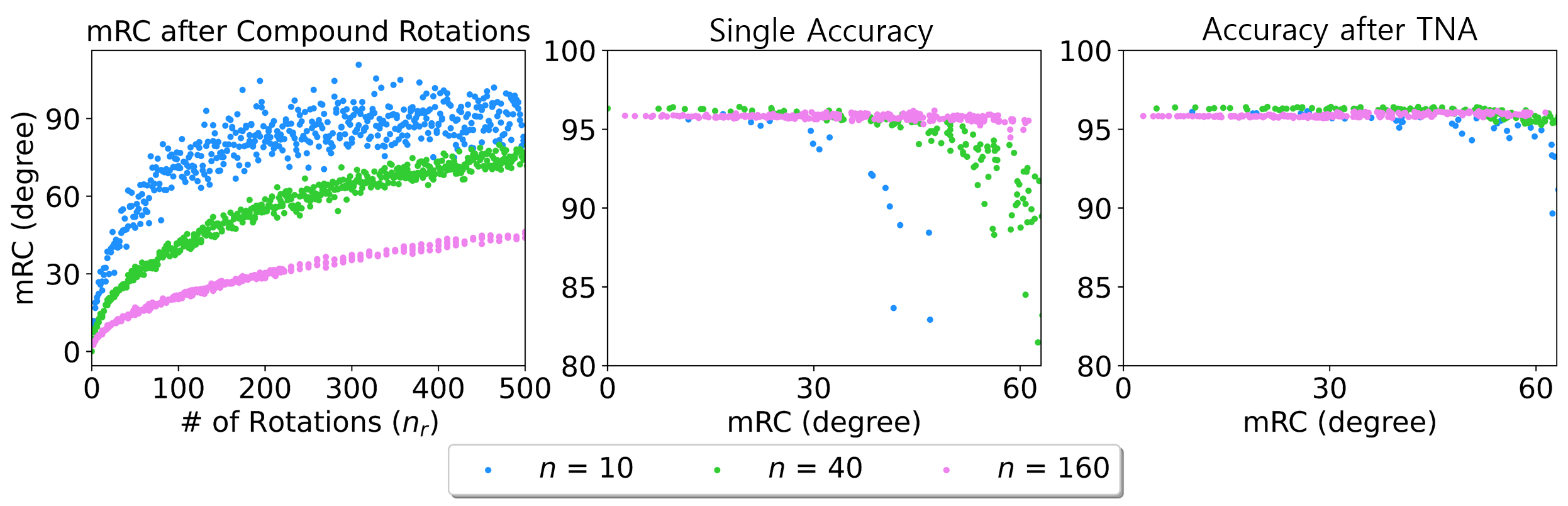}
     \caption{Ablation study of when the dimension of penultimate feature $n$ decrease. The experiment is done on CIFAR-10 dataset, with WideResNet28x10. Here when $n$ is set to lower values, the mRC tend to have larger deviation for a certain $n_r$, making it harder for the user to generate for a certain mRC. Thus the accuracy interpolation performance degrades after \textsc{Tna}.}
     \label{fig:dimension_ablation_plot_supp}
 \end{figure}

2) The penultimate feature space is high-dimensional : The assumption $\Delta_{\mathbf{z},i}=\arccos{(\cos\psi_i \cos\theta)} - \psi_i$ is associated with the high-dimensionality assumption.
As mentioned in the "Blessings of the dimensionality" section in \cref{sec: discussion}, the essence of our approach is to apply a transformation to the class vector based on intensity, adjusting the correlation with the penultimate feature. When a high-intensity transform is applied, the correlation between the penultimate feature and the class vector decreases, approaching 90 degrees, as indicated by the near orthogonal theorem. If the dimensionality decreases, the effectiveness of the algorithm may diminish. To verify this, we present additional ablation studies in Tab.\ref{tab:num_n_differ} and Fig.\ref{fig:dimension_ablation_plot_supp}. First, in Tab.\ref{tab:num_n_differ}, we varied the dimension $n$ of the last linear layer while training WideResNet28x10\cite{zagoruyko2016wide}, MobileNetV2\cite{Sandler2018MobileNetV2}, and EfficientNetB0\cite{li2019efficient} on CIFAR10 and CIFAR100 datasets. The additional linear layer is added in between the feature extractor and the final linear layer for the experiment. We measured the ECE, and it is evident that the effectiveness of the algorithm decreases as $n$ decreases for both datasets. Next, in Fig.\ref{fig:dimension_ablation_plot_supp}, we measured the utility of the \textsc{Tilt} algorithm as $n$ decreases. The left figure, similar to Fig.~\ref{fig:plot_compund_rotations}, measures mRC by increasing $n_r$, and it shows that the narrower the $n$, the wider the range of mRC generated by a specific $n_r$. Therefore, it indicates the need for delicate adjustment of mRC for optimal performance of the proposed algorithm. The middle and right figures measure the decrease in accuracy for different mRCs in situations with a single tilted weight (middle) and \textsc{Tna} weight $n_e=10$ (right). Again, it confirms that as $n$ decreases, accuracy performance degradation occurs at smaller angles.

\begin{table}[h!]
  \begin{center}
    \caption{Expected Calibration Error in percentage(\%) for two datasets with respect to change in the number of dimension($n$) in the penultimate feature(pf)., CIFAR100 (left) and CIFAR10(right). None stands for unscaled, \textsc{Tna} (Tilt and Average), TS(Temperature Scaling), and TS+\textsc{Tna}. The number aside to the architecture, with the parentheses denote the original dimension. Bold for the best calibrated, underline for the second best. Lower the better. }
    \label{tab:num_n_differ}
    \centering 

    \begin{tabular}{c|c|cc|cc||cc|cc}
    \toprule
      &&\multicolumn{4}{c}{\textbf{CIFAR100}} &\multicolumn{4}{c}{\textbf{CIFAR10}}\\ \cline{3-10}
      \textbf{Model}&$n$ & None & \textsc{Tna} & TS & +\textsc{Tna} & None & \textsc{Tna} & TS & +\textsc{Tna}\\
      \midrule
      \multirow{4}{*}{WRN28x10 (640)} & 10 & 11.05 & 11.04 &  \underline{1.41} & \textbf{1.40} & 1.73 & 1.48 & 0.7 & 0.48 \\
      &40 & 7.24 &  \underline{2.21} & 3.10 & \textbf{2.18} & 1.78 & 0.55 & 0.58 & 0.54 \\
      &160 & 7.25 &  \underline{2.66} & 2.84 & \textbf{2.05} & 2.20 & 0.60 & 0.83 & 0.54 \\
      &640 & 5.91 &  \underline{4.11} & 4.69 & \textbf{4.55} & 1.66 & 0.66 & 0.84 & 0.78 \\
      \hline
      \multirow{4}{*}{\makecell{MNV2(1280)}} & 20 & 10.23 & 10.23 & \underline{1.85} & \textbf{1.80} & 3.44 & 2.78 & \underline{0.79} & \textbf{0.65}\\
       & 80 & 10.92 & 1.74 & \underline{1.48} & \textbf{1.31} & 3.09 & \underline{0.58} & 0.69 & \textbf{0.53} \\
       &  320 & 11.51 & \underline{1.26} & 1.52 & \textbf{1.19} & 3.51 & \underline{0.66} & 0.68 & \textbf{0.54} \\
       & 1280 & 10.03 & \underline{1.54} & 1.82 & \textbf{1.41} & 3.49 & \textbf{0.78} & 1.06 & \underline{0.81} \\
       \midrule
       \multirow{4}{*}{EffB0(320)} & 40 & 10.9 & 1.63 & \underline{1.59} & \textbf{1.37} & 2.74 & 1.15 & \underline{0.96} & \textbf{0.86} \\
       & 80 & 10.7 & 1.52 & \textbf{1.32} & \textbf{1.32} & 3.67 & \underline{0.85} & 0.95 & \textbf{0.71} \\
       &  160 & 10.23 & \underline{1.24} & 1.29 & \textbf{1.01} & 3.24 & \underline{0.73} & 0.81 & \textbf{0.66} \\
       &  320 & 14.8 & \underline{1.56} & 1.77 & \textbf{1.34} & 4.65 & \textbf{0.82} & 1.12 & \underline{0.95} \\
      \bottomrule
    \end{tabular}
  \end{center}
\end{table}

\section{Proof of Theorem~\ref{thm:ndim_mainthm_lem_main}}
\label{app: proof}

\begin{proof} 
For the purpose of descriptive convenience, let $u$ be the direction vector of class vector$\mathbf{w}_i$ be $u$, and direction vector of penultimate vector $\mathbf{z}$ be v and $\psi_i$ be $\psi$. Without loss of generality, let $u$ be a unit vector pointing at the north pole of the first dimension. $u=[1,0,...,0]$, and $v=[v_1,v_2,...,v_n]$ Then the vector $v$ lies in the space where $\angle{(u,v)}=\arccos(u \cdot v)=\psi$. That is, 
\begin{equation}
u \cdot v = v_1 = \cos \psi, v_2^2 + \cdots v_n^2 = 1- \cos^2 \psi = \sin^2 \psi.
\end{equation}

Without loss of generality, we can also suppose that the plane $u,v$ lies in the subspace spanned by two basis vectors $[1,0, \cdots,0]$ and $[0,1,0,\cdots,0]$, which means $v_1=\cos\psi, v_2 = \sin\psi$, $v_3 = v_4 = \cdots =0$.

The rotated vector of $v$, $R_{\theta}(v)=[t_1,t_2, \cdots, t_n]$ will lie in the trace of the intersection of the surface of the n-dimension hypersphere where $t_1^2+t_2^2+ \cdots +t_n^2 = 1$ and $v \cdot R_{\theta}(v) = \cos \theta$. 
$t_1 \cos\psi+t_2 \sin\psi = \cos\theta$. As a function of $t_1$, $ \left\{ t_2, t_3, \cdots t_n \right\}$ has the trace of,
\begin{equation}
\begin{aligned}
t_2 &= \frac{\cos\theta - t_1 \cos\psi}{\sin\psi}, \\
t_3^2+t_4^2+ \cdots + t_n^2 &= 1-t_1^2-t_2^2 \\ &= 1-{t_1}^2 -\left(\frac{\cos\theta-t_1\cos\psi}{\sin\psi}\right)^2,
\end{aligned}
\end{equation}
that is when $t_1$ is fixed, $\{t_3, \cdots,t_n\}$ will lie in the surface of the $n-2$ dimensional hypersphere of radius $\sqrt{1-{t_1}^2 -\left(\frac{\cos\theta-t_1\cos\psi}{\sin\psi}\right)^2}$.

\begin{equation}
\begin{aligned}
1-{t_1}^2& -\left(\frac{\cos\theta-t_1\cos\psi}{\sin\psi}\right)^2 \\& = 1-t_1^2 - \frac{\cos^2\theta-2t_1\cos\psi\cos\theta+t_1^2\cos^2\psi}{\sin^2 \psi}\\
&=-\frac{1}{\sin^2\psi}t_1^2 + \frac{2\cos\psi\cos\theta}{\sin^2 \psi}t_1 - \frac{\cos^2 \theta}{\sin^2 \psi} +1\\
&=-\frac{1}{\sin^2\psi} \left(t_1^2 - 2\cos\psi\cos\theta + \cos^2\psi \cos^2\theta \right) \\
&\qquad\qquad\qquad+\frac{\cos^2\psi\cos^2\theta}{\sin^2\psi} - \frac{\cos^2\theta}{\sin^2\psi}+ \frac{\sin^2\psi}{\sin^2\psi}\\
&=-\frac{1}{\sin^2\psi} \left(t_1-\cos\psi\cos\theta \right)^2 +\sin^2 \theta,
\end{aligned}
\end{equation}
and yields its maximum value when $t_1 = \cos\psi\cos\theta$. 
Also 
\begin{equation}
\begin{aligned}
t_3^2+t_4^2+ &\cdots + t_n^2 = 1-{t_1}^2 -t_2^2 \\
&=-\frac{1}{\sin^2\psi} \left(t_1-\cos\psi\cos\theta \right)^2 +\sin^2 \theta \geq 0
\end{aligned}
\end{equation}
gives the fact that $ \cos(\psi+\theta) \leq t_1 \leq \cos(\psi-\theta) $, because the quadratic function is unimodal. The angle $\angle(u,R_{\theta}(v)) = \arccos{u \cdot v} = \arccos{t_1}$ is a strictly decreasing function, to find the mode it is sufficient to find the $t_1$, 
\begin{equation}
\mathbb{M}[\Delta] =\arccos\left({\arg\max_{t_1}{\sqrt{1-{t_1}^2 -\left(\frac{\cos\theta-t_1\cos\psi}{\sin\psi}\right)^2}}}\right ) - \psi,
\end{equation}
which is proved to yield its maximum value when $t_1 = \cos\psi\cos\theta$.

Additionally,  
\begin{equation}
    \begin{aligned}
        \frac{\partial}{\partial\psi}&\left(\arccos{\left( \cos\psi \cos\theta \right)} - \psi \right) \\&= -\frac{1}{\sqrt{1-(\cos\psi \cos\theta)^2}}\cdot (-sin\psi) \cdot (\cos{\theta})-1 \\
        &=\frac{\sin\psi \cos\theta}{\sqrt{1-\cos^2{\psi} \cos^2 \theta}}-1 \\
        &<\frac{\sin\psi}{\sqrt{1-\cos^2{\psi}}}-1 = \frac{\sin\psi}{\sin\psi}-1 = 0,
     \end{aligned}
\end{equation}
and the $\mathbb{M}[\Delta_{\mathbf{z},i}]=arccos(\cos\psi\cos\theta) - \psi$ is the strictly decreasing function of $\psi$.

\end{proof}

\section{Real-life Implications}

\begin{figure}[h]
\label{fig:real_life}
    \centering
    \includegraphics[width=0.95\linewidth]{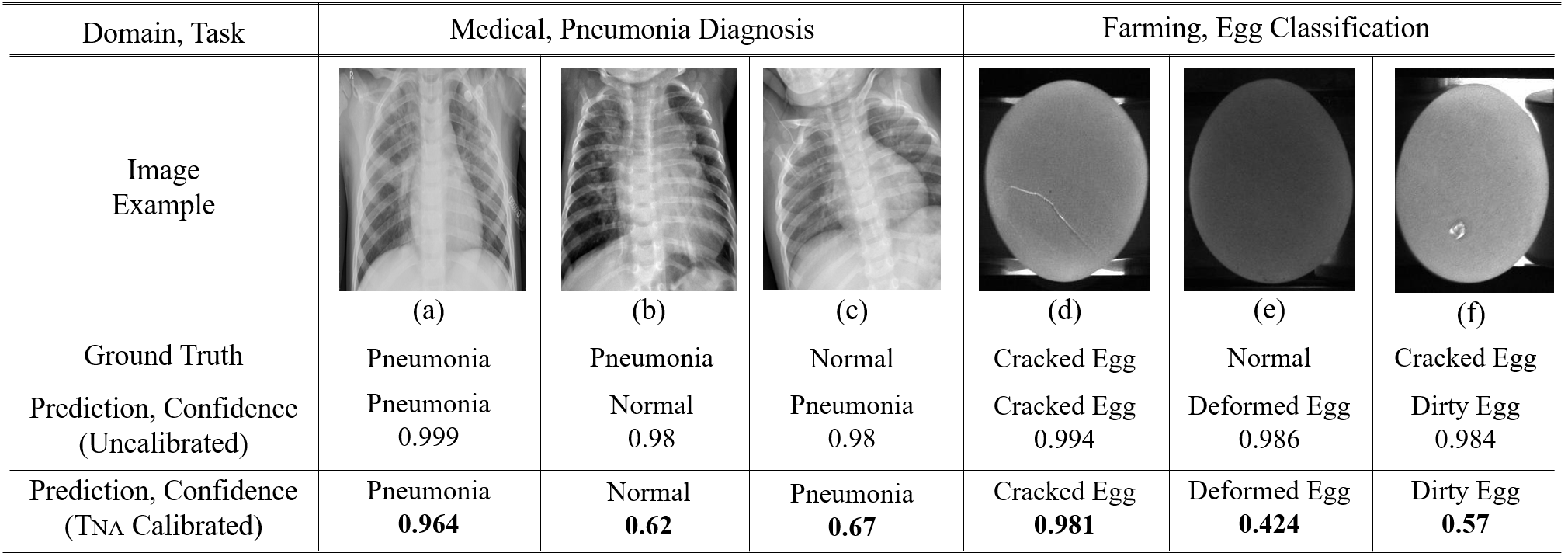}
    \caption{Real-life examples of confidence calibration.}
\end{figure}

\textbf{(Pneumonia Diagnosis Case.)} First, consider a binary classification problem, specifically diagnosing pneumonia using only chest images \cite{Kermany2018OCT}((a), (b), (c)). In case (a), where the individual is diagnosed with pneumonia, the confidence in pneumonia infection exceeds 0.9, indicating a high confidence level in the correct prediction. However, in case (b), a high confidence of 0.98 was assigned to the prediction of not having pneumonia. After calibration, the confidence level decreases to 0.62. This lower confidence level in binary classification may serve as a signal of uncertainty in the decision. False negatives and false positives can be life-threatening in the medical domain.

\textbf{(Egg Classification Case.)} Consider the task of discerning fresh eggs \cite{EggData}(d),(e),(f). This task involves classifying eggs based on their photos into different categories such as Normal, Cracked Egg, Deformed Egg, and Dirty Egg. We may utilize the confidence of the calibrated model as an indicator of uncertainty (e.g., using a threshold of 0.5). In case (d), an egg with a crack is correctly classified as a cracked egg after calibration with high confidence. In case (e), where a saleable egg is misclassified as a Deformed Egg with miscalibrated high confidence, and can be discarded. In case (f), a cracked egg can be misclassified as a dirty egg, where some of the dirty eggs can be sold after washing, but eggs with cracks cannot be sold after washing . This could lead to significant losses for the farm.

\section{Experimental Details}

\begin{table} [!htbp]
\centering
\begin{tabular}{l|cccc}
\toprule
                & $lr_{init}$                 & epochs                  & wd  &  batch size                     \\ \hline
WideResNet28x10 & 0.1                      & 300                     & 5e-4      &128               \\
GoogleNet  & 0.1                      & 300                     & 5e-4      &128          \\
MobileNetV2     & {0.1} & {250} & {3e-4}  & 128\\
PreResNet110    & 0.1                      & 300                     & 5e-4      &128       \\ \hline

\end{tabular}

\caption{Training Details of trained weights for SVHN.}
\label{app : tab_SVHN_details}
\end{table}

\begin{table} [!htbp]
\centering
\begin{tabular}{l|cccc}
\toprule
                & $lr_{init}$                 & epochs                  & wd  &  batch size                     \\ \hline
WideResNet28x10 & 0.1                      & 300                     & 5e-4      &128               \\
GoogleNet  & 0.1                      & 300                     & 3e-4         & 128   \\
MobileNetV2     & {0.01} & {250} & {3e-4}  & 128\\
PreResNet110    & {0.1}  & {300} & {3e-4}  & 128\\ \hline

\end{tabular}

\caption{Training Details of trained weights for CIFAR10 and CIFAR100.}
\label{app : tab_CIFAR_details}
\end{table}

We list the detailed information to conduct the experiments that are included in the main paper. 

\textbf{Model Family.} The alternative weights are generated on WideResNet28x10\citep{zagoruyko2016wide}, GoogleNet\citep{szegedy2015going}, PreResNet110\citep{he2016deep}, and MobileNetV2\citep{Sandler2018MobileNetV2} for CIFARs, and ResNet-50\citep{he2016deep}, and DenseNet169\citep{huang2017densely}, ViT-L/16, and ViT-H/14\citep{dosovitskiy2021an} . To train the model for the CIFAR10, and CIFAR100, we followed the standard training scheme and evaluation code borrowed from the pioneering work of \cite{ashukha2020pitfalls}, with the training information written in Tab.~\ref{app : tab_SVHN_details}, \ref{app : tab_CIFAR_details} , and \ref{app : tab_CIFARweights_details}.
 For the ImageNet-1k dataset, the trained weights are borrowed from Timm\citep{rw2019timm} and Torchvision \citep{Marcel2010Torchvision}. We fix the number of ensembles $n_e = 10$ for all the models. 

 \textbf{Training Details.} Most of the settings to train SVHN/CIFAR10/CIFAR100 networks are taken from \cite{ashukha2020pitfalls}. Specifically, we used Momentum-SGD with batch size of 128, momentum 0.9. The batch size is fixed to 256. The learning rate scheduler follows the work of \cite{Garipov2018Loss} and is specified as, 

 \begin{equation}
lr(i) \equiv \begin{cases} lr_{init} &  \frac{i}{epochs} \in [0,0.5] \\
lr_{init}(1.0-0.99(\frac{\frac{i}{epochs}-0.5}{0.4})) & \frac{i}{epochs} \in [0.5,0.9] \\
lr_{init} \times 0.01 & otherwise  
\end{cases}
 \end{equation}

 The information of the trained models is listed in Table.~\ref{app : tab_ImageNet_details}.

\begin{table*}[!htbp]

\scalebox{0.8}{
\begin{tabular}{l|ccccccc}
\toprule
               &  Acc. (SVHN) & Acc. (CF10)      & Acc. (CF100)                & Image Size &    Interpolation     & Mean & Std \\ \hline
WideResNet28x10 & 97.03 & 96.42   & 81.25     & 32 &  padding=4    & [0.4914, 0.4822, 0.4465]    &  [0.2023, 0.1994, 0.2010]      \\
GoogleNet   & 96.32 &  95.24 & 79.41   & 32 & padding=4   & [0.4914, 0.4822, 0.4465]   &  [0.2023, 0.1994, 0.2010]       \\
MobileNetV2  &  95.64 & 92.58  & 73.41  & 32 &  padding=4  & [0.4914, 0.4822, 0.4465]   & [0.2023, 0.1994, 0.2010]    \\
PreResNet110  &  96.565  & 94.71   & 77.74  & 32 &  padding=4  & [0.4914, 0.4822, 0.4465]   & [0.2023, 0.1994, 0.2010]    \\
 
 \hline 

\end{tabular}
}

\caption{The information of the trained models for CIFAR10 and CIFAR100 datasets. We list the top-1, top-5 accuracy, and train / test transformations for the images.}
\label{app : tab_CIFARweights_details}
\end{table*}

\begin{table*}[!htbp]
\scalebox{0.9}{
\begin{tabular}{l|ccccccc}
\toprule
                & Top-1                 & Top-5                 & Image Size &    Crop Percentage & Interpolation     & Mean & Std \\ \hline
ResNet-50 & 76.11                     & 92.862  & 224 & 0.875 & bicubic    & [0.485, 0.456, 0.406]    &  [0.229, 0.224, 0.225]      \\
DenseNet-169     & 75.6            & 92.806  & 224 & 0.875 & bicubic  & [0.485, 0.456, 0.406]    &  [0.229, 0.224, 0.225]   \\
  ViT-L/16    & 84.24  & 97.818 & 224 &  0.9 & bicubic & [0.485, 0.456, 0.406]    &  [0.229, 0.224, 0.225]   \\
 ViT-H/14    & 85.708     & 97.73  & 224 & 0.875 & bicubic & [0.485, 0.456, 0.406]    &  [0.229, 0.224, 0.225]  \\ 
 \hline 

\end{tabular}
}
\caption{The information of the trained models for ImageNet datasets. We list the top-1, top-5 accuracy, and train / test transformations for the images.}
\label{app : tab_ImageNet_details}
\end{table*}

\textbf{Hyperparameters.} Generally we set the number of ensembles $n_e=10$ for all the results, and $\alpha=5, \beta=1$ for the hyperparameters for beta distribution.  $\theta_s$ is set to 0.9rad if not stated. The search interval of rotation number $n_t$ is set to 50.  We set three different $\theta_s$ for all the models, of 0.5, 1.0, 1.5 . The value of $\theta_s$ for Fig.~\ref{fig:plot_compund_rotations} is specified in the figure. $n$, the dimension of the penultimate feature($pf$), is a pre-determined value from the model, with values corresponding to each model. 

\end{document}